\documentclass{bmvc2k}
\pdfoutput=1

\usepackage{amsfonts}
\usepackage{amsmath}
\usepackage{amssymb}
\usepackage{booktabs}
\usepackage{enumerate}
\usepackage{graphicx}
\usepackage{makecell}
\usepackage{microtype}
\usepackage{multirow}
\usepackage{nicefrac}
\usepackage{xcolor}

\title{Weak-shot Semantic Segmentation by Transferring Semantic Affinity and Boundary}

\addauthor{Siyuan Zhou}{ssluvble@sjtu.edu.cn}{1}
\addauthor{Li Niu}{ustcnewly@sjtu.edu.cn}{1*}
\addauthor{Jianlou Si}{sijianlou@sensetime.com}{2}
\addauthor{Chen Qian}{qianchen@sensetime.com}{2}
\addauthor{Liqing Zhang}{zhang-lq@cs.sjtu.edu.cn}{1}

\addinstitution{
 MoE Key Lab of Artificial Intelligence,\\
 Shanghai Jiao Tong University\\
 Shanghai, China
}
\addinstitution{
 SenseTime Research,\\
 SenseTime\\
 Beijing, China
}
\runninghead{Siyuan Zhou}{Weak-shot Semantic Segmentation by RETAB}

\begin{document}

\maketitle

\begin{abstract}
Weakly-supervised semantic segmentation (WSSS) with image-level labels has been widely studied to relieve the annotation burden of the traditional segmentation task. In this paper, we show that existing fully-annotated base categories can help segment objects of novel categories with only image-level labels, even if base categories and novel categories have no overlap. We refer to this task as weak-shot semantic segmentation, which could also be treated as WSSS with auxiliary fully-annotated categories. Recent advanced WSSS methods usually obtain class activation maps (CAMs) and refine them by affinity propagation. Based on the observation that semantic affinity and boundary are class-agnostic, we propose a method under the WSSS framework to transfer semantic affinity and boundary from base to novel categories. As a result, we find that pixel-level annotation of base categories can facilitate affinity learning and propagation, leading to higher-quality CAMs of novel categories. Extensive experiments on PASCAL VOC 2012 dataset prove that our method significantly outperforms WSSS baselines on novel categories.
\end{abstract}

\section{Introduction}\label{sec:introduction}

Semantic segmentation~\cite{bertasius2017convolutional, chen2017deeplab, lin2016efficient, long2015fully, noh2015learning, qi2016hierarchically, yu2015multi, hu2018learning} is fundamental in computer vision and has been greatly advanced through the rapid development of deep learning techniques. Traditional fully-supervised segmentation heavily relies on expensive pixel-level (full) annotations. To solve this issue, segmentation paradigms requiring fewer or weaker annotations have gradually attracted research attention, like weakly-/semi-supervised segmentation~\cite{wei2018revisiting, papandreou2015weakly, souly2017semi, hong2015decoupled} and one-/few-shot segmentation~\cite{shaban2017one, zhang2020sg, dong2018few, rakelly2018few, hu2019attention}. However, these paradigms have limitations in practical applications. Semi-supervised/few-shot segmentation can not handle new categories with no pixel-level annotations. Weakly-supervised semantic segmentation (WSSS) leverages more accessible weak annotations, but the performance gap between WSSS and fully-supervised segmentation is still non-negligible.

\begin{figure*}
	\centering
	\includegraphics[width=\linewidth]{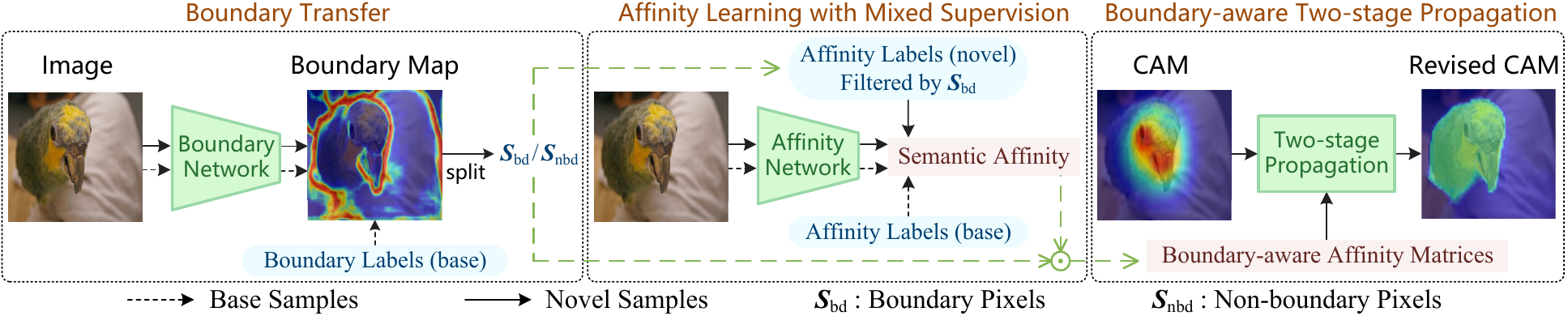}
	\caption{Overview of our RETAB: (\romannumeral1) train a boundary network to transfer semantic boundary from base to novel categories, (\romannumeral2) train an affinity network to learn semantic affinity in a mixed-supervised manner, and (\romannumeral3) perform boundary-aware two-stage propagation to revise CAMs based on the learned semantic boundary and affinity. Details can be found in Section~\ref{subsec:RETAB}.}
	\label{fig:architecture}
\end{figure*}

In this work, we propose a new learning paradigm called weak-shot semantic segmentation. Assume that we have a fully-annotated segmentation dataset containing training samples of only base categories (marked as base samples). We are also provided with extra weakly-annotated training samples containing objects of base or novel categories (marked as novel samples). Therefore, our problem can be considered as weakly-supervised segmentation with an auxiliary fully-annotated dataset, but the set of well-labeled categories is limited. At test time, our objective is to segment images where both base and novel categories may exist. The critical problem in our task is transferring class-agnostic knowledge from base categories to novel ones to enhance the segmentation performance of novel categories. Considering different types of weak annotation,
we focus on image-level labels in this work since it is a popular research direction in WSSS~\cite{chen2020weakly, fan2020learning, wei2017object, wei2018revisiting, araslanov2020single, zhang2020reliability}. That is, base samples have pixel-level labels while novel samples only have image-level labels in our problem. 

We draw inspiration from classical WSSS methods to start our work. Until recently, some advanced WSSS methods are based on Class Activation Map (CAM)~\cite{zhou2016learning} which can effectively localize discriminative parts of objects by training a classification network with image-level labels. The typical WSSS framework usually obtains CAM as the initial response, expands the response region to acquire pseudo labels, and uses pseudo labels to train a segmentation network. Among WSSS methods under this framework, PSA~\cite{ahn2018learning} is a representative one with two main steps: learn semantic affinities between pair-wise pixels within local neighborhoods on the feature map (\emph{i.e.}, affinity learning), and generate a transition matrix to perform random walk~\cite{lovasz1993random} on CAMs (\emph{i.e.}, affinity-based propagation).

In this paper, we design our method under the typical WSSS framework and focus on better expanding the initial response. With the assumption that semantic affinities and boundaries are class-agnostic, we attempt to transfer these two types of information from base to novel categories. Our method is called \textbf{R}esponse \textbf{E}xpansion by \textbf{T}ransferring semantic \textbf{A}ffinity and \textbf{B}oundary (RETAB), which contains an affinity learning step and an affinity-based propagation step. In the affinity learning step, our goal is to design an affinity network that learns semantic affinities from ground-truth labels of base samples and CAMs of novel samples. Inspired by prior works on instance segmentation~\cite{ke2021deep, fan2020commonality}, boundary knowledge could function as an effective tool to assist with affinity learning. However, CAMs are noisy on semantic boundaries, leading to imprecise semantic affinities for novel categories. To solve this problem, we first consider boundary (\emph{resp.}, non-boundary) pixels as unconfident (\emph{resp.}, confident) pixels~\cite{chen2020weakly}. Then we propose to train a boundary network supervised by base samples. The predicted boundaries of novel samples are used to split apart CAMs into boundary regions and non-boundary regions. To filter out noisy supervisions in CAMs of novel samples, we only consider non-boundary regions when we train the affinity network.

In the affinity-based propagation step, we propose a two-stage propagation strategy to revise CAMs. In the first stage, random walk is restricted within the non-boundary regions. Pixels with dominant category labels in the confident regions will be propagated to fit object shapes. In the second stage, unconfident pixels in the boundary regions are propagated under the guidance of confident pixels in the non-boundary regions. Hopefully, confident pixels can regulate the random walk of unconfident ones and facilitate propagation on object boundaries. After propagation, pseudo segmentation labels are obtained based on the revised CAMs (revised responses). Finally, we train a segmentation network under the mixed supervisions from ground-truth labels of base samples and pseudo labels of novel samples, which is the only network used in the inference stage. In summary, our main contributions are:
\begin{itemize}
	\item We study a novel paradigm called weak-shot semantic segmentation that utilizes full annotations of base categories to benefit segmenting objects of novel categories with only image-level labels.
	\item We propose a simple yet effective method called RETAB to transfer class-agnostic semantic affinity and semantic boundary from base to novel categories under the typical WSSS framework, together with \emph{a novel boundary-aware two-stage propagation strategy}. Our method can be integrated into any WSSS method under this framework.
	\item The effectiveness of RETAB is verified on PASCAL VOC 2012 dataset~\cite{everingham2010pascal}. RETAB significantly outperforms WSSS baselines and naive weak-shot segmentation baselines on novel categories.
\end{itemize}

\section{Related Works}\label{sec:related}

\noindent\textbf{Weakly-supervised Semantic Segmentation:}
Weakly-supervised semantic segmentation (WSSS)~\cite{singh2017hide, fan2018associating, sun2020mining, kwak2017weakly} has attracted considerable interest because weak annotations are conveniently available. Most advanced WSSS methods with image-level labels~\cite{chang2020weakly, ahn2018learning, zhang2020causal, wang2020self, lee2021anti, zhang2021complementary} are based on the class activation map (CAM)~\cite{zhou2016learning} obtained from a classification network. ``Seed, expand, and constrain'', three principles proposed by SEC~\cite{kolesnikov2016seed}, are followed by many WSSS works. Some of them work on improving the seed or initial response~\cite{lee2019ficklenet, chang2020weakly, wang2020self}. The other works followed the \emph{coarse-to-fine} strategy to expand or propagate responses~\cite{huang2018weakly,wang2018weakly,ahn2018learning,zhang2020splitting}. Compared with WSSS, our proposed weak-shot segmentation takes advantage of full annotations in existing datasets, and can be considered as WSSS with an auxiliary fully-annotated dataset containing limited well-labeled categories. In this work, our RETAB is realized under the typical WSSS framework and focuses on the response expansion problem.

\noindent\textbf{Few-/Zero-shot Semantic Segmentation:}
Few-shot semantic segmentation~\cite{shaban2017one, zhang2020sg, dong2018few, hu2019attention} and zero-shot semantic segmentation~\cite{bucher2019zero, gu2020context} have been studied by plenty of works in recent years. They both divide the whole category set into base categories and novel categories. Few-shot semantic segmentation assumes that only a few training images are available for each novel category, but pixel-level annotations are still required for novel categories. Zero-shot semantic segmentation relies on category-level semantic representations that are often weak and ambiguous, so the performance is far from satisfactory. Likewise, our proposed weak-shot semantic segmentation has a split of base categories and novel categories. Differently, we provide novel categories with weak annotations that are easily accessible and useful for learning segmentation models. 

\noindent\textbf{Semi-supervised Semantic Segmentation:}
Typically, semi-supervised semantic segmentation~\cite{hung2018adversarial, ouali2020semi, hong2015decoupled, ibrahim2020semi} addresses the issue of utilizing a set of well-labeled images to enhance the segmentation quality for another set of unlabeled images~\cite{hung2018adversarial, ouali2020semi} or weakly-labeled images~\cite{hong2015decoupled, ibrahim2020semi}. In this work, our proposed weak-shot semantic segmentation follows a similar idea to separate training samples into two sets with different annotation levels. Differently, our task further splits categories into base ones and novel ones, which involves cross-category knowledge transfer.

\noindent\textbf{Weak-shot Learning:}
Actually, the weak-shot learning paradigm, \emph{i.e.}, full annotations for base categories and weak annotations for novel categories, has been studied in image classification~\cite{chen2020weakshot}, object detection~\cite{liu2021mixed, zhong2020boosting, chen2020cross, li2018mixed}, and instance segmentation~\cite{hu2018learning, kuo2019shapemask, zhou2020learning, fan2020commonality, biertimpel2021prior}. Weak-shot classification~\cite{chen2020weakshot} supposes that base categories have clean image labels and novel categories only have noisy ones. Weak-shot detection~\cite{liu2021mixed}, also called mixed-supervised~\cite{li2018mixed} or cross-supervised~\cite{chen2020cross} detection, requires that base categories have box-level annotations while novel ones only have image-level labels. Weak-shot instance segmentation, usually called partially-supervised instance segmentation, utilizes mask annotations of base categories and only bounding boxes of novel ones. The abovementioned methods generally transfer class-agnostic target (\emph{e.g.}, similarity, objectness) or learn the mapping from weak annotation to full annotation. To the best of our knowledge, weak-shot semantic segmentation has not been explored. Compared with weak-shot classification/detection,  we focus on the more challenging segmentation task. Compared with weak-shot instance segmentation, we only utilize higher-available image-level labels instead of stronger boxes. 

\begin{figure*}
	\centering
	\includegraphics[width=\linewidth]{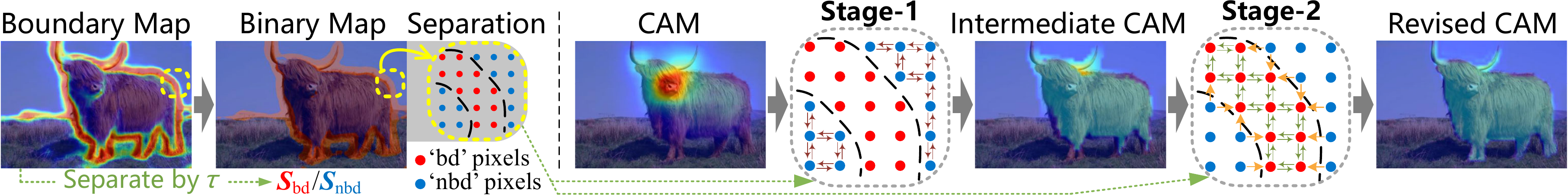}
	\caption{Illustration of boundary-aware two-stage propagation. The predicted boundary map is separated into boundary (`bd') pixels $\mathcal{S}_{\rm bd}$ and non-boundary (`nbd') pixels $\mathcal{S}_{\rm nbd}$. In the first stage only `nbd' pixels are propagated, while in the second stage `nbd' pixels guide the propagation of `bd' pixels. See Section~\ref{para:twostage} for more details.}
	\label{fig:two-stage}
\end{figure*}

\section{Methodology}\label{sec:methodaology}

In weak-shot semantic segmentation, we have base categories $\mathcal{C}^b$ and novel categories $\mathcal{C}^n$, satisfying $\mathcal{C} = \mathcal{C}^b\cup \mathcal{C}^n$ and $\mathcal{C}^b\cap \mathcal{C}^n=\emptyset$. We assume that background $bg$ belongs to base categories, \emph{i.e.}, $bg\in \mathcal{C}^b$. Training data contain $\mathcal{N}^b$ base samples and $\mathcal{N}^n$ novel samples with no intersection. Base samples only contain base categories, whereas novel samples may contain base or novel categories. We provide pixel-level labels for base samples and only image-level labels for novel samples. We use $c_i^*$ to denote the ground-truth segmentation label of the $i$-th pixel on the feature map for any base sample. Next, we will first take a glance at PSA~\cite{ahn2018learning}, which represents the typical WSSS framework. Then, we describe our RETAB to transfer class-agnostic semantic affinity and boundary under this framework, which is outlined in Figure~\ref{fig:architecture}. Finally, we discuss the training of the segmentation network.

\subsection{Review of CAM and PSA}\label{subsec:psa}

As mentioned in Section~\ref{sec:introduction}, the typical WSSS framework~\cite{ahn2018learning, chang2020weakly, wang2020self} adopts CAM as the initial response, which is obtained by training a classification network with image-level labels. CAM $M_l$ of a typical category $l\in\mathcal{C}$ highlights the discriminative regions of this category. A group of WSSS methods design models to augment CAM~\cite{chang2020weakly, wang2020self}, which could also function as the initial response. Another group of WSSS methods attempts to design effective algorithms to expand responses~\cite{huang2018weakly,wang2018weakly,ahn2018learning}, in which PSA~\cite{ahn2018learning} is a representative one. Specifically, PSA designs an AffinityNet to learn feature map $f^{\rm aff}$, based on which pair-wise semantic affinity $\hat{a}_{ij}$ between pixel $i$ and $j$ is calculated by
$\hat{a}_{ij} = \exp \big\{ -{\left \| f^{\rm aff}(x_i,y_i)-f^{\rm aff}(x_j,y_j) \right\|}_1 \big\}$. In the training stage, PSA only considers pixel pairs in neighbor set $\mathcal{P}$:
\begin{eqnarray}\label{eqn:coord_aff}
	\mathcal{P}=\big\{(i,j) \mid d\big( (x_i,y_i),(x_j,y_j) \big)<\gamma,\;i\neq j\big\} \,.
\end{eqnarray}
where $\gamma$ is a search radius and $d(\cdot,\cdot)$ represents Euclidean distance. The network is supervised by affinity labels $a_{ij}$ obtained from CAMs (refer to~\cite{ahn2018learning} for details), where $a_{ij}=1$ for intra-category pairs and $a_{ij}=0$ for inter-category pairs. The trained AffinityNet enforces the predicted semantic affinity $\hat{a}_{ij}$ to be close to $1$ if pixel $i$ and $j$ belong to the same category, and close to $0$ otherwise. Predicted $\hat{a}_{ij}$ form the affinity matrix $\hat{A}$, with which random walk is performed on each $M_l$ to obtain the revised response of category $l$. Next, we describe our RETAB that adapts response expansion for weak-shot semantic segmentation.

\subsection{Pipeline of RETAB}\label{subsec:RETAB}

\paragraph{Boundary Transfer.}\label{para:boundary}

We first introduce transferring semantic boundaries from base categories to novel ones because this will facilitate our two major steps to expand responses. As shown in Figure~\ref{fig:architecture}, we train a boundary network to predict a boundary map from the input image. We use $\hat{p}_i$ to represent the predicted boundary probability (after $\mathrm{Sigmoid}$) for pixel $i$ on the feature map. During training, the boundary network is only supervised with base samples. Boundary labels $b^*_i$ for base samples are derived from segmentation labels, and we denote $b^*_i=1/0$ for boundary/non-boundary pixels. In training, the model tends to suppress responses in boundary regions, which is caused by the unbalanced boundary, foreground, and background pixels in the training samples. To address it, we split pixels into three subsets: boundary pixels $\mathcal{S}_{\rm bd}=\{i \mid b^*_i=1\}$, non-boundary foreground pixels $\mathcal{S}_{\rm fg}=\{i \mid b^*_i=0, c_i^*\in\mathcal{C}^b \backslash \{bg\} \}$, and non-boundary background pixels $\mathcal{S}_{\rm bg}=\{i \mid b^*_i=0,c_i^*=bg\}$. Cross-entropy classification losses are applied to three sets and merged to form the total loss of the boundary network:
\begin{eqnarray}\label{eqn:boundary_loss}
	\mathcal{L}_{B}=-\sum_{i\in \mathcal{S}_{\rm bd}}{\frac{\log{(\hat{p}_i})}{|\mathcal{S}_{\rm bd}|}}-\frac{1}{2}\sum_{i\in \mathcal{S}_{\rm fg}}{\frac{\log{(1-\hat{p}_i})}{|\mathcal{S}_{\rm fg}|}}\nonumber-\frac{1}{2}\sum_{i\in \mathcal{S}_{\rm bg}}{\frac{\log{(1-\hat{p}_i})}{|\mathcal{S}_{\rm bg}|}} \,.
\end{eqnarray}
We expect the class-agnostic boundary knowledge embedded in the trained boundary model to be transferred to novel categories. Therefore, we perform boundary prediction on novel samples. To facilitate affinity learning and affinity-based propagation, we pre-set a threshold $\tau$ to divide the boundary prediction for each novel sample into two parts: a boundary region $\hat{\mathcal{S}}_{\rm bd}=\{i \mid \hat{p}_i\geqslant\tau\}$ and a non-boundary region $\hat{\mathcal{S}}_{\rm nbd}=\{i \mid \hat{p}_i<\tau\}$, as shown in Figure~\ref{fig:two-stage}.

\paragraph{Affinity Learning with Mixed Supervision.}\label{para:trainAffnet}

In PSA, AffinityNet is supervised by coarse affinity labels $a_{ij}$ obtained from CAMs, which are imprecise because CAMs only highlight discriminative parts. We similarly implement an affinity network to predict pair-wise semantic affinities $\hat{a}_{ij}$ within local neighborhoods. Differently, we adopt more robust affinity labels $a^*_{ij}$ by utilizing fully-annotated base samples and weakly-annotated novel samples to train the network in a mixed-supervised manner (see Figure~\ref{fig:architecture}). The improvement comes from two aspects. Firstly, affinity labels of base samples are purified through pixel-level annotations. This helps the model learn better semantic affinities and facilitates novel categories by assuming that semantic affinities are class-agnostic. For base samples, affinity labels $a^*_{ij}$ in $\mathcal{P}$ are obtained from the segmentation label, \emph{i.e.}, $a^*_{ij}=1$ when $c_i^*=c_j^*$, and $a^*_{ij}=0$ otherwise. Secondly, pixel-level labels for novel samples are not given, so we can only obtain coarse affinity labels from CAMs. Our idea is to leverage predicted boundaries to purify the supervision. Since inaccurate pixels usually occur on CAM boundaries~\cite{chen2020weakly}, we only use affinity labels in non-boundary regions. Specifically, we narrow the set $\mathcal{P}$ in (\ref{eqn:coord_aff}) to a cleaner set $\mathcal{P}_{\rm nbd}$ by filtering out pixel pairs in boundary regions:
$\mathcal{P}_{\rm nbd} = \big\{(i,j) \mid d\big( (x_i,y_i),(x_j,y_j) \big) < \gamma,\; i\neq j \mathrm{\;\, and \;\,} i,j\notin\hat{\mathcal{S}}_{\rm bd}  \big\} $.
For novel samples, our affinity labels $a^*_{ij}$ in $\mathcal{P}_{\rm nbd}$ are obtained in the same way as $a_{ij}$ (see Section \ref{subsec:psa}). By filtering out noisy affinity labels, the model can learn more accurate affinities from novel samples. Details of the cross-entropy loss to optimize the affinity network can be found in PSA. The loss form remains unchanged and is suitable for all samples by replacing $a_{ij}$ with $a^*_{ij}$ for base samples and replacing $\mathcal{P}$ with $\mathcal{P}_{\rm nbd}$ for novel ones. 

\paragraph{Boundary-aware Two-stage Propagation.}\label{para:twostage}

After training the affinity network, we predict the semantic affinity $\hat{a}_{ij}$ between pixel $i$ and $j$ within the entire neighbor set $\mathcal{P}$. Similar to \cite{ahn2018learning}, we perform random walk on CAMs based on $\hat{a}_{ij}$ which represents the transition probability that pixel $i$ should be propagated to pixel $j$. Distinctive from propagating on the whole CAM for a single stage in the classical random walk, we adopt a boundary-aware two-stage propagation strategy that concurrently utilizes semantic affinities and boundaries (see Figure~\ref{fig:two-stage}). In the first stage, the random walk is restricted within non-boundary regions to prevent the disturbance from unconfident pixels on the boundaries, in which case the boundary regions are analogous to isolation belts. Under this design, the pixels with dominant category labels will be propagated to other pixels to fit object shapes and complete the masks. Restricting the propagation area can be implemented by setting certain entries in the affinity matrix to zero, leading to a sparse affinity matrix $\hat{A}^{(1)}$ as follows:
\begin{eqnarray}\label{eqn:affmat_step1}
	{\hat{A}}_{\,ij}^{(1)}=
	\left\{
	\begin{aligned}
		\hat{a}_{ij},\quad &\forall\;i\neq j \mathrm{\;\; s.t. \;\,} i,j\in\hat{\mathcal{S}}_{\rm nbd},\;\, (i,j)\in\mathcal{P},\\
		1\;\;,\quad &\forall\;i=j \,,\\
		0\;\;,\quad &{\rm otherwise} \,.
	\end{aligned}
	\right. 
\end{eqnarray}
In the second stage, we hope that confident non-boundary pixels propagated in the first stage can regulate the random walk process of boundary pixels, whereas unconfident boundary pixels are not allowed to affect confident ones, so the sparse affinity matrix $\hat{A}^{(2)}$ should be
\begin{eqnarray}\label{eqn:affmat_step2}
	{\hat{A}}_{\,ij}^{(2)}=
	\left\{
	\begin{aligned}
		\hat{a}_{ij},\quad &\forall\;i\neq j \mathrm{\;\; s.t. \;\,} j\in\hat{\mathcal{S}}_{\rm bd},\;\, (i,j)\in\mathcal{P}, \\
		1\;\;,\quad &\forall\;i=j \,,\\
		0\;\;,\quad &{\rm otherwise} \,.
	\end{aligned}
	\right. 
\end{eqnarray}
We can observe that $\hat{A}^{(1)}$ is bidirectional, but $\hat{A}^{(2)}$ contains unidirectional components, indicating the critical difference between two-stage propagation and the original random walk with only one bidirectional $\hat{A}$. Our two-stage propagation is applied on CAM $M_l$ by firstly using $\hat{A}^{(1)}$ and secondly using $\hat{A}^{(2)}$ as the affinity matrix in the random walk process. We strictly follow~\cite{ahn2018learning} to perform random walk with affinity matrix without tuning parameters. In detail, we first generate transition probability matrices corresponding to $\hat{A}^{(1)}$ and $\hat{A}^{(2)}$. Then, the random walk in each stage is accomplished by iteratively multiplying the corresponding transition matrix to $M_l$ until a predefined number of iterations is reached. We refer to the CAMs after boundary-aware two-stage propagation as revised CAMs (or revised responses).

\begin{figure*}
	\centering
	\includegraphics[width=0.74\linewidth]{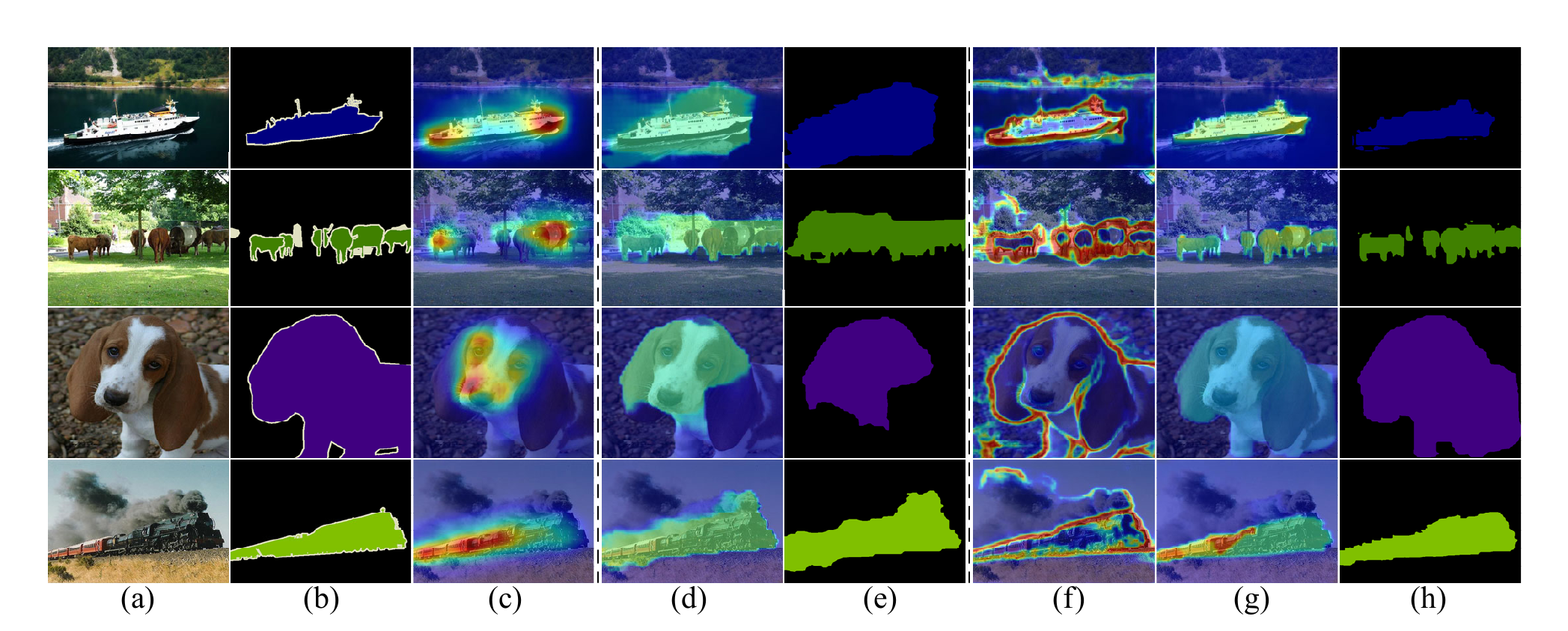}
	\caption{Visualized pseudo labels on VOC12 \emph{train} set. Examples from top to bottom belong to novel samples in fold 0,1,2,3, respectively. (a) image. (b) GT. (c) CAM. (d)(e) CAM+RW and pesudo labels. (f) boundary prediction. (g)(h) CAM+RETAB and pesudo labels.}
	\label{fig:cam}
\end{figure*}

\subsection{Mixed-supervised Segmentation}\label{subsec:mixedSegmentation}

Since revised responses have smaller resolutions than input images, we first up-sample them to the original sizes using bilinear interpolation. Then, we apply $\mathrm{argmax}$ on the category channel of the concatenated revised responses to obtain pseudo segmentation labels of novel samples. Finally, we can train any segmentation network in a fully-supervised manner using mixed supervisions from ground-truth labels of base samples and pseudo labels of novel samples. During inference, the segmentation network takes in a test image to predict categories within $\mathcal{C}^b\cup \mathcal{C}^n$ for each pixel because test images may contain either base or novel categories.

\section{Experiments}\label{sec:experiment}

\subsection{Experimental Setting}\label{subsec:settings}

\paragraph{Datasets and Evaluation Metrics.}\label{para:dataset}

Following most WSSS works, we conduct experiments on PASCAL VOC 2012 dataset~\cite{everingham2010pascal} with 21 classes, including 20 foreground object classes and a background class. The official dataset includes 1464 training images, 1449 validation images, and 1456 test images. Following common practice in semantic segmentation, we adopt an augmented training set with 10582 images from SBD~\cite{hariharan2011semantic}. Following the category split rule in PASCAL-$\mathbf{5^i}$~\cite{shaban2017one}, which is commonly used in few-shot segmentation~\cite{zhang2020sg, dong2018few, hu2019attention, wang2019panet}, we evenly divide the 20 foreground categories into four folds. Categories in each fold are regarded as 5 novel categories, and the remaining categories (including background) are regarded as 16 base categories. We further divide 10582 training samples into base samples and novel samples for each fold. Images containing only base categories are included as base samples, whereas those containing at least one novel category are deemed novel samples. More separation details for categories and samples are left to supplementary. For each fold, we retain full annotations of base samples and only image-level labels for novel samples. During inference, we use the official validation set and test set to verify the segmentation performance. We adopt Intersection-over-Union (IoU) as the evaluation metric. Due to the specific characteristics of our task, we calculate mean IoU on all categories, base categories, and novel categories, which are referred to as all-mIoU, base-mIoU, and novel-mIoU, respectively.

\paragraph{Implementation Details.}\label{para:detail}

The boundary network is based on ResNet38~\cite{wu2019wider}. We concatenate feature maps in shallow layers and deep layers from three stages with $1\times 1$ convolutions. The concatenated feature map is followed by another $1\times 1$ convolution with the output dimension being one and a $\mathrm{Sigmoid}$ layer. The network structure of our affinity network is the same as AffinityNet in PSA~\cite{ahn2018learning}, which is also based on ResNet38. For both boundary network and affinity network, we augment training images with color jittering, random cropping ($448\times 448$), and horizontal flip. Both networks are trained with batch size 8 for eight epochs. SGD~\cite{bottou2010large} optimizer is adopted with weight decay of $5e^{-4}$. The learning rate is initialized as $0.001$ (\emph{resp.}, $0.01$) for the boundary network (\emph{resp.}, affinity network) and decreases following the polynomial policy $lr_{\rm iter}=lr_{\rm initial}(1-\frac{\rm iter}{\rm max\_iter})^{\alpha}$ with $\alpha=0.9$. We strictly follow WSSS baselines~\cite{ahn2018learning, wang2020self, zhang2021complementary} to adopt DeepLab~\cite{chen2014semantic} with ResNet38~\cite{wu2019wider} backbone pretrained on ImageNet~\cite{deng2009imagenet} as the segmentation network. The training schemes (data augmentation, batch size, training epoch, optimizer, learning rate, and other parameters) remain unchanged for segmentation. All experiments are conducted on 2 NVIDIA RTX 2080Ti GPUs with PyTorch.

\paragraph{Baselines.}\label{para:baseline}

Our RETAB can be incorporated into any WSSS baseline under the typical framework with three steps: 1) obtain CAM as the initial response, 2) propagate response to acquire pseudo labels, and 3) train a segmentation network. We incorporate our method into PSA~\cite{ahn2018learning}, SEAM~\cite{wang2020self}, and CPN~\cite{zhang2021complementary}. They all use random walk with AffinityNet (\emph{abbr.} RW~\cite{ahn2018learning, ahn2019weakly}) as the second step, and the same segmentation network in the third step. The difference lie in the first step. PSA directly uses ``CAM'' proposed in~\cite{zhou2016learning} as the initial response, while SEAM and CPN design augmented responses based on their proposed architectures. For ease of representation, we use ``CAM+RW'', ``SEAM+RW'', and ``CPN+RW'' to denote the overall pipelines of baselines. For fairness, when integrating our method into each baseline, we use the same initial response and segmentation network as this baseline, resulting in our ``CAM+RETAB'', ``SEAM+RETAB'', and ``CPN+RETAB''.

\begin{table*}[t]
	\centering
	\begin{tabular}{p{32mm}p{3.5mm}<{\centering}p{3.5mm}<{\centering}p{3.5mm}<{\centering}p{3.5mm}<{\centering}p{3.5mm}<{\centering}p{3.5mm}<{\centering}p{3.5mm}<{\centering}p{3.5mm}<{\centering}p{3.5mm}<{\centering}p{3.5mm}<{\centering}p{3.5mm}<{\centering}p{3.5mm}<{\centering}}
		\toprule
		\multirow{2}*{Method} & \multicolumn{3}{c}{fold 0} & \multicolumn{3}{c}{fold 1} & \multicolumn{3}{c}{fold 2} & \multicolumn{3}{c}{fold 3} \\
		& $\mathcal{C}$ & $\mathcal{C}^b$ & $\mathcal{C}^n$ & $\mathcal{C}$ & $\mathcal{C}^b$ & $\mathcal{C}^n$ & $\mathcal{C}$ & $\mathcal{C}^b$ & $\mathcal{C}^n$ & $\mathcal{C}$ & $\mathcal{C}^b$ & $\mathcal{C}^n$ \\
		\midrule
		SSDD~\cite{shimoda2019self} & 65.5 & 67.6 & 58.8 & 65.5 & 64.5 & 68.7 & 65.5 & 63.5 & 72.1 & 65.5 & 68.0 & 57.7 \\
		BES~\cite{chen2020weakly} & 66.6 & 68.8 & 59.6 & 66.6 & 64.9 & 71.9 & 66.6 & 64.7 & 72.6 & 66.6 & 69.3 & 57.8 \\
		SvM~\cite{zhang2020splitting} & 66.7 & 67.5 & 64.2 & 66.7 & 65.8 & 69.7 & 66.7 & 65.6 & 70.6 & 66.7 & 69.6 & 57.7 \\
		\midrule
		CAM+RW~\cite{ahn2018learning} & 63.7 & 65.4 & 58.1 & 63.7 & 63.7 & 63.8 & 63.7 & 61.4 & 71.0 & 63.7 & 65.8 & 56.8 \\
		CAM+RW(seggt) & 73.8 & 78.5 & 58.6 & 74.8 & \textbf{76.5} & 69.5 & 73.7 & 74.4 & 71.5 & 73.9 & 79.2 & 56.9 \\
		CAM+RW(affgt+seggt) & 75.2 & 78.7 & 64.0 & 75.3 & \textbf{76.5} & 71.5 & 74.6 & 75.2 & 72.7 & 74.1 & \textbf{79.3} & 57.5 \\
		CAM+RETAB & \textbf{76.3} & \textbf{78.8} & \textbf{68.0} & \textbf{76.0} & 76.1 & \textbf{75.9} & \textbf{75.4} & \textbf{75.4} & \textbf{75.6} & \textbf{74.8} & 79.2 & \textbf{60.8} \\
		\midrule
		SEAM+RW~\cite{wang2020self} & 65.7 & 67.8 & 59.0 & 65.7 & 64.7 & 68.9 & 65.7 & 63.7 & 72.3 & 65.7 & 68.2 & 57.9 \\
		SEAM+RW(seggt) & 74.0 & 78.7 & 59.1 & 74.5 & 75.6 & 71.1 & 73.5 & 73.3 & 74.0 & 73.7 & 78.1 & 59.6 \\
		SEAM+RW(affgt+seggt) & 74.9 & \textbf{78.9} & 62.1 & 75.2 & 76.4 & 71.4 & 74.3 & 74.3 & 74.3 & 74.2 & 78.7 & 59.8 \\
		SEAM+RETAB & \textbf{75.5} & \textbf{78.9} & \textbf{64.6} & \textbf{76.0} & \textbf{76.6} & \textbf{74.0} & \textbf{75.1} & \textbf{75.0} & \textbf{75.6} & \textbf{74.8} & \textbf{79.0} & \textbf{61.5} \\
		\midrule
		CPN+RW~\cite{zhang2021complementary} & 68.5 & 70.7 & 61.5 & 68.5 & 66.8 & 73.8 & 68.5 & 66.6 & 74.5 & 68.5 & 71.2 & 59.7 \\
		CPN+RW(seggt) & 74.7 & 78.6 & 62.4 & 75.5 & 76.0 & 74.0 & 75.5 & 75.7 & 74.8 & 74.4 & 78.8 & 60.2 \\
		CPN+RW(affgt+seggt) & 76.1 & 79.0 & 66.8 & 76.5 & \textbf{76.8} & 75.7 & 75.6 & 75.6 & 75.5 & 74.9 & \textbf{79.3} & 60.7 \\
		CPN+RETAB & \textbf{76.6} & \textbf{79.1} & \textbf{68.8} & \textbf{76.7} & 76.7 & \textbf{76.7} & \textbf{75.9} & \textbf{75.8} & \textbf{76.2} & \textbf{75.3} & \textbf{79.3} & \textbf{62.4} \\
		\midrule
		Fully Oracle & 77.9 & 79.1 & 74.4 & 77.9 & 77.7 & 78.7 & 77.9 & 76.4 & 82.9 & 77.9 & 79.6 & 72.6 \\
		\bottomrule
	\end{tabular}
	\caption{Comparison of segmentation performance on VOC12 \emph{test} set. Columns marked by $\mathcal{C}$/$\mathcal{C}^b$/$\mathcal{C}^n$ represents all-/base-/novel-mIoU. We have five groups of methods. The first group includes some popular WSSS methods. The second / third / fourth group represents the WSSS baseline PSA / SEAM / CPN, together with our created weak-shot segmentation baselines (seggt and affgt+seggt), and our RETAB. The fifth group is a fully-supervised method. Except the first group, all experiments use DeepLab~\cite{chen2014semantic} with ResNet38~\cite{wu2019wider} backbone.}
	\label{table:seg}
\end{table*}

\paragraph{Hyper-parameters.}\label{para:hyperparameter}

During boundary prediction, we set the threshold $\tau$ as $0.5$ via cross-validation (see supplementary for detailed discussion). For the other hyper-parameters (search radius $\gamma=5$, the parameters to generate affinity labels $a^*_{ij}$ for novel samples and the parameters for random walk in each stage of our two-stage propagation), we use the default values in WSSS baseline PSA~\cite{ahn2018learning}, SEAM~\cite{wang2020self}, and CPN~\cite{zhang2021complementary} without further tuning.


\begin{figure*}
	\centering
	\includegraphics[width=0.9\linewidth]{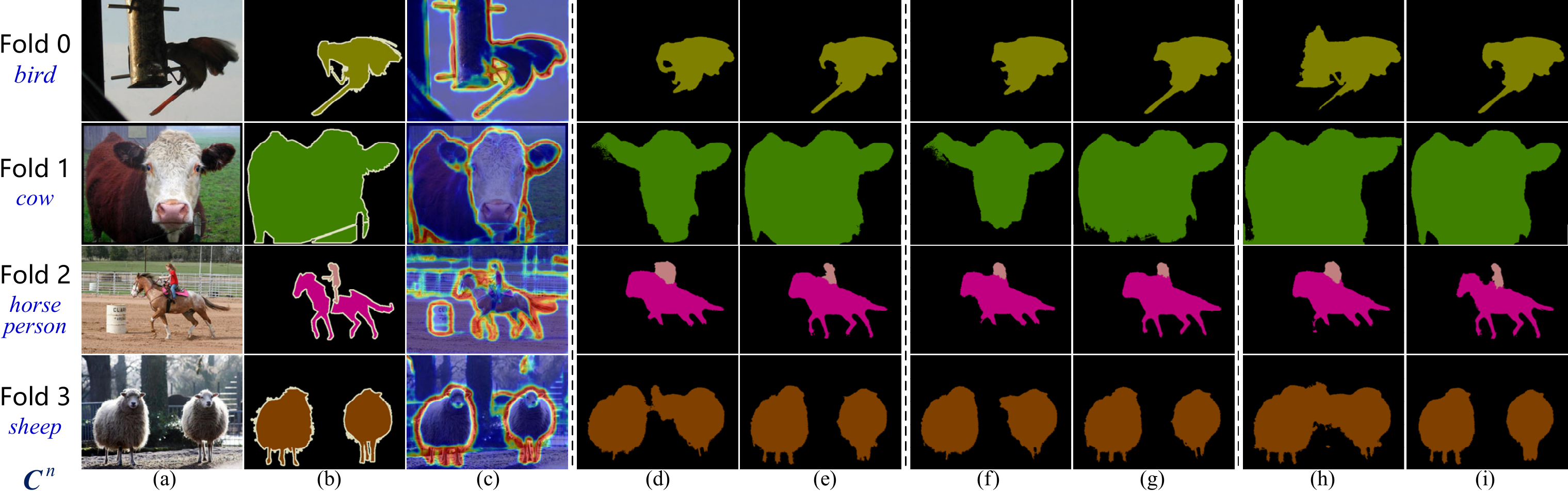}
	\caption{Visualized segmentation results of novel categories on VOC12 \emph{val} set. Examples from top to bottom contain objects of novel categories (marked as blue on the left side) in fold 0,1,2,3, respectively. (a) image. (b) GT label. (c) boundary prediction. (d) CAM+RW. (e) CAM+RETAB. (f) SEAM+RW. (g) SEAM+RETAB. (h) CPN+RW. (i) CPN+RETAB.}
	\label{fig:seg}
\end{figure*}

\subsection{Evaluation on Pseudo Segmentation Labels}\label{subsec:EXPpseudolabel}

We evaluate pseudo segmentation labels by assessing CAMs before/after propagation on VOC12 \emph{train} set. Figure~\ref{fig:cam} visualizes the pseudo segmentation labels of some novel samples by comparing ``CAM+RW'' with ``CAM+RETAB''. As shown, pseudo labels generated by RETAB better adapt to object boundaries than the classical random walk, demonstrating RETAB's transferability of semantic affinity and boundary from base categories to novel ones. More quantitative analyses and visualizations results can be found in the supplementary.

\subsection{Evaluation on Segmentation Performance}\label{subsec:EXPsegmentation}

Naturally, the fully-supervised (\emph{resp.,} weakly-supervised) setting serves as the upper (\emph{resp.,} lower) bound of our weak-shot setting. In this subsection, we compare the final segmentation results of our method with the upper/lower bound. Quantitative results on VOC12 \emph{test} set are summarized in Table~\ref{table:seg} (see supplementary for results on the \emph{val} set). ``CAM+RW'', ``SEAM+RW'' and ``CPN+RW'' represent three WSSS baselines we use. We correspondingly create three naive weak-shot segmentation baselines, denoted by ``(seggt)''. The simple modification is to train the final segmentation step with the mixed supervisions of ground-truth segmentation labels for base samples and pseudo segmentation labels generated by WSSS baselines for novel samples. Also, we create three augmented weak-shot segmentation baselines, denoted by ``(affgt+ seggt)''. Based on naive baselines, these augmented baselines further utilize ground-truth affinity labels of base samples when training the affinity network. Our methods are represented by ``CAM+RETAB'', ``SEAM+RETAB'' and ``CPN+RETAB''. ``Fully Oracle'' is a reproduced fully-supervised baseline. As illustrated in Table~\ref{table:seg}, our RETAB consistently and significantly outperforms WSSS baselines and weak-shot baselines for novel categories on all four folds, verifying the effectiveness of knowledge transfer across categories. Compared with ``Fully Oracle'', our method can recover $97.7\%\sim100\%$  of its bound for base categories and $83.7\%\sim97.5\%$ of its bound for novel categories, implying that RETAB successfully narrows the gap between weakly-supervised and fully-supervised segmentation. Figure~\ref{fig:seg} shows some visualizations on segmenting objects of novel categories with different methods (see supplementary for more visualizations). Our RETAB works better in recovering object shapes, especially in boundary regions.

\subsection{Ablation Studies and Generalization to Other Settings}\label{subsec:ablation_generalization}

We carefully analyze the functionality of boundary transfer, semantic affinity, and boundary-aware two-stage propagation, respectively. Different parts in our method are verified to demonstrate the source of improvement for the segmentation results. A significance test is also included. For more practical application, we generalize our weak-shot semantic segmentation task to two settings: 1) generalization to potential novel categories in the background, and 2) generalization to fewer fully-annotated training data. 
We observe that our RETAB could generalize to discover potential cues of novel categories in the background of base samples with a simple follow-up self-training step. We also observe that RETAB can use a small proportion of base samples to facilitate a large number of novel samples.
These detailed results can be found in the supplementary.

\section{Conclusion}\label{sec:conclusion}

This paper has proposed a novel paradigm called weak-shot semantic segmentation that utilizes pixel-level annotations of base categories to improve the segmentation performance on novel categories with only image-level labels. Under the typical WSSS framework, a simple yet effective method called RETAB is developed to expand response regions by transferring class-agnostic semantic affinity and boundary. 
Our work provides a simple yet effective baseline to promote future research on weak-shot semantic segmentation.

\section*{Acknowledgements}

The work was supported by the Shanghai Municipal Science and Technology Major/Key Project, China (2021SHZDZX0102, 20511100300) and  National Natural Science Foundation of China (Grant No. 61902247).

\bibliography{ref}
\end{document}


\maketitle

In this document, we provide additional materials to support our main submission. In Section~\ref{subsec:app_basicdataset}, we introduce the basic splits of PASCAL VOC 2012 dataset~\cite{everingham2010pascal} used in the main paper. All the experiments in Section~\ref{subsec:app_hyperparameter},~\ref{subsec:app_boundary},~\ref{subsec:app_affinity},~\ref{subsec:app_propagation},~\ref{subsec:app_improve},~\ref{subsec:app_pseudores},~\ref{subsec:app_segres},~\ref{subsec:app_vispseudo} and~\ref{subsec:app_visseg} depend on these basic splits. In Section~\ref{subsec:app_hyperparameter}, we analyze the hyper-parameters in our model. In Section~\ref{subsec:app_boundary},~\ref{subsec:app_affinity} and~\ref{subsec:app_propagation}, we perform ablation studies to verify the functionalities of different parts in our method. In Section~\ref{subsec:app_improve}, we analyze the source of improvement for the final segmentation results. 
In Section~\ref{subsec:app_pseudores},~\ref{subsec:app_segres},~\ref{subsec:app_vispseudo} and~\ref{subsec:app_visseg}, we supplement experiments in the main paper with more quantitative and qualitative results of pseudo segmentation labels and final segmentation results. In Section~\ref{subsec:app_potential} and~\ref{subsec:app_unbalanced}, we then extend the basic splits to more general ones to deal with additional problems in weak-shot semantic segmentation. Finally, we discuss the limitations and future works in Section~\ref{subsec:app_limitation}.

\section{Basic Splits of PASCAL VOC 2012 Dataset}\label{subsec:app_basicdataset}

As mentioned in Section 4.1 in the main paper,  we follow the category splits in PASCAL-$\mathbf{5^i}$~\cite{shaban2017one} to form four basic folds, \emph{i.e.}, fold 0, fold 1, fold 2, and fold 3. In fold $i$ ($0\leqslant i\leqslant 3$), categories with indices $5i+1,5i+2,5i+3,5i+4,5i+5$ are regarded as $5$ novel categories and the remaining categories (including ``background'') are regarded as 16 base categories. The detailed split of base categories and novel categories for each fold is shown in Table~\ref{table:app_basicdataset}.

We use the augmented training set ($trainaug$ set) with 10582 images from SBD~\cite{hariharan2011semantic} as the complete training set for our experiments conducted on PASCAL VOC 2012 dataset. Based on the abovementioned category splits, we further divide 10582 training samples into base samples and novel samples for each fold. The images containing only base categories are included as base samples, whereas those containing at least one novel category are deemed novel samples. The total count of base samples and novel samples in each fold are listed in Table~\ref{table:app_basecount} for both the $trainaug$ set and the official $train$ set.

Following recent WSSS researches~\cite{ahn2018learning, chen2020weakly, wang2020self, zhang2020splitting}, we adopt $trainaug$ set for model training, $train$ set for evaluating CAMs or pseudo labels, and $val$ set or $test$ set for evaluating segmentation results.

\begin{table}[t]
  \centering
  \begin{tabular}{ccccc}
	\toprule
	& \multicolumn{2}{c}{$trainaug$} & \multicolumn{2}{c}{$train$} \\
	\textbf{fold} & $\mathcal{N}^b$ & $\mathcal{N}^n$ & $\mathcal{N}^b$ & $\mathcal{N}^n$ \\
	\cmidrule(r){2-3}\cmidrule(r){4-5}
	\textbf{0} & 7746 & 2836 & 1051 & 413  \\
    \textbf{1} & 6978 & 3604 & 950 & 514  \\
    \textbf{2} & 5040 & 5542 & 800 & 664  \\
    \textbf{3} & 8437 & 2145 & 1091 & 373  \\
	\bottomrule
  \end{tabular}
  \caption{The number of base samples and novel samples in fold $i$ ($0\leqslant i\leqslant 3$). $\mathcal{N}^b$ and $\mathcal{N}^n$ represent the number of base samples and  novel samples, respectively.}
  \label{table:app_basecount}
\end{table}

\begin{table}[t]
    \centering
    \begin{tabular}{cccccc}
      \toprule
      \textbf{fold} & sample & acc. & prec. & recall & ${\rm F}_1$-score\\
      \cmidrule(r){3-6}
      \textbf{0} & base & 0.831 & 0.594 & 0.777 & 0.612 \\
      \textbf{0} & novel & 0.854 & 0.597 & 0.774 & 0.622 \\
      \midrule
      \textbf{1} & base & 0.846 & 0.593 & 0.777 & 0.615 \\
      \textbf{1} & novel & 0.836 & 0.594 & 0.743 & 0.614 \\
      \bottomrule
    \end{tabular}
    \caption{Analysis of boundary prediction results on PASCAL VOC 2012 \emph{train} set. Base samples and novel samples in fold 0 and fold 1 are evaluated separately.}
    \label{table:app_boundary}
\end{table}

\begin{table*}[t]
  \scriptsize
  \centering
  \begin{tabular}{p{5.8mm}p{0mm}p{1mm}p{1.2mm}p{1.2mm}p{1.2mm}p{2.4mm}p{0.3mm}p{0.2mm}p{0mm}p{1.8mm}p{0.9mm}p{1.8mm}p{0.6mm}p{2.1mm}p{1.3mm}p{3.4mm}p{2.3mm}p{2.5mm}p{1.7mm}p{2.1mm}p{1.4mm}}
	\toprule
	\textbf{class} & bg & aero & bike & bird & boat & bottle & bus & car & cat & chair & cow & table & dog & horse & mbk & person & plant & sheep & sofa & train & tv \\
	\textbf{index} & 0 & 1 & 2 & 3 & 4 & 5 & 6 & 7 & 8 & 9 & 10 & 11 & 12 & 13 & 14 & 15 & 16 & 17 & 18 & 19 & 20 \\
	\midrule
	\textbf{fold 0} & $b$ & $n$ & $n$ & $n$ & $n$ & $n$ & $b$ & $b$ & $b$ & $b$ & $b$ & $b$ & $b$ & $b$ & $b$ & $b$ & $b$ & $b$ & $b$ & $b$ & $b$ \\
    \textbf{fold 1} & $b$ & $b$ & $b$ & $b$ & $b$ & $b$& $n$ & $n$ & $n$ & $n$ & $n$ & $b$ & $b$ & $b$ & $b$ & $b$ & $b$ & $b$ & $b$ & $b$ & $b$ \\
    \textbf{fold 2} & $b$ & $b$ & $b$ & $b$ & $b$ & $b$ & $b$ & $b$ & $b$ & $b$ & $b$ & $n$ & $n$ & $n$ & $n$ & $n$ & $b$ & $b$ & $b$ & $b$ & $b$ \\
    \textbf{fold 3} & $b$ & $b$ & $b$ & $b$ & $b$ & $b$ & $b$ & $b$ & $b$ & $b$ & $b$ & $b$ & $b$ & $b$ & $b$ & $b$ & $n$ & $n$ & $n$ & $n$ & $n$ \\
	\bottomrule
  \end{tabular}
  \caption{The basic splits of 21 categories for PASCAL VOC 2012 dataset. In each fold, `$b$' denotes a base category and `$n$' denotes a novel category. $bg$ is the background category.}
  \label{table:app_basicdataset}
\end{table*}

\section{Analysis of Hyper-parameter}\label{subsec:app_hyperparameter}

Recall that we use a threshold $\tau$ to divide the boundary prediction results into boundary regions and non-boundary regions, as discussed in Section 3.2 in the main paper. To investigate the impact of $\tau$, we conduct experiments by trying different $\tau$ in the range of $[0.1,0.9]$ on all four folds. Table~\ref{table:app_threshold} summarizes the \emph{train} set mIoUs of revised responses, which utilize CAM~\cite{zhou2016learning} as the initial response and our proposed RETAB as the propagation strategy.  We can observe that most high mIoUs occur when $\tau$ lies in the range of $[0.3, 0.7]$. In our implementation, we consistently set $\tau=0.5$ in all experiments.

\begin{table*}[t]
  \centering
  \begin{tabular}{p{7mm}<{\centering}p{5mm}<{\centering}p{5mm}<{\centering}p{5mm}<{\centering}p{5mm}<{\centering}p{5mm}<{\centering}p{5mm}<{\centering}p{5mm}<{\centering}p{5mm}<{\centering}p{5mm}<{\centering}p{5mm}<{\centering}p{5mm}<{\centering}p{5mm}<{\centering}}
	\toprule
	& \multicolumn{3}{c}{fold 0} & \multicolumn{3}{c}{fold 1} & \multicolumn{3}{c}{fold 2} & \multicolumn{3}{c}{fold 3} \\
	$\tau$ & $\mathcal{C}$ & $\mathcal{C}^{b}$ & $\mathcal{C}^{n}$ & $\mathcal{C}$ & $\mathcal{C}^{b}$ & $\mathcal{C}^{n}$ & $\mathcal{C}$ & $\mathcal{C}^{b}$ & $\mathcal{C}^{n}$ & $\mathcal{C}$ & $\mathcal{C}^{b}$ & $\mathcal{C}^{n}$ \\
	\midrule
	0.1 & 69.0 & 71.7 & 60.1 & 68.6 & 68.3 & 69.7 & 68.6 & 68.1 & 70.3 & 69.0 & 70.8 & 63.3 \\
	0.2 & 70.3 & 72.8 & 62.3 & 70.3 & 70.0 & 71.0 & 70.1 & 69.4 & 72.3 & 70.1 & 71.9 & \textbf{64.1} \\
	0.3 & 70.9 & 73.4 & \textbf{62.8} & 70.8 & 70.5 & \textbf{71.9} & 70.7 & 70.0 & 72.9 & \textbf{70.3} & 72.3 & \textbf{64.1} \\
	0.4 & 71.0 & 73.7 & 62.6 & 71.1 & 70.9 & 71.6 & 70.8 & 70.1 & 73.0 & 70.2 & 72.4 & 63.4 \\
	0.5 & \textbf{71.2} & 74.0 & 62.5 & 71.3 & 71.2 & 71.6 & \textbf{70.9} & 70.2 & \textbf{73.3} & 70.1 & 72.4 & 62.8 \\
	0.6 & 71.1 & 74.0 & 62.1 & 71.3 & 71.3 & 71.4 & \textbf{70.9} & 70.2 & 73.2 & 70.1 & \textbf{72.5} & 62.3 \\
	0.7 & \textbf{71.2} & \textbf{74.2} & 61.8 & \textbf{71.4} & \textbf{71.5} & 71.2 & \textbf{70.9} & \textbf{70.4} & 72.7 & 69.9 & 72.4 & 61.9 \\
	0.8 & 71.1 & 74.1 & 61.7 & 71.3 & 71.4 & 71.0 & \textbf{70.9} & \textbf{70.4} & 72.6 & 69.8 & 72.4 & 61.5 \\
	0.9 & 71.1 & 74.1 & 61.5 & 71.3 & 71.4 & 70.8 & 70.8 & 70.3 & 72.5 & 69.4 & 72.3 & 61.2 \\
	\bottomrule
  \end{tabular}
  \caption{Different choices of hyper-parameter $\tau$ on four folds. The mIoU evaluation of revised responses for ``CAM+RETAB'' is conducted on PASCAL VOC 2012 \emph{train} set.}
  \label{table:app_threshold}
\end{table*}

\begin{figure*}[t]
  \centering
  \includegraphics[width=0.88\linewidth]{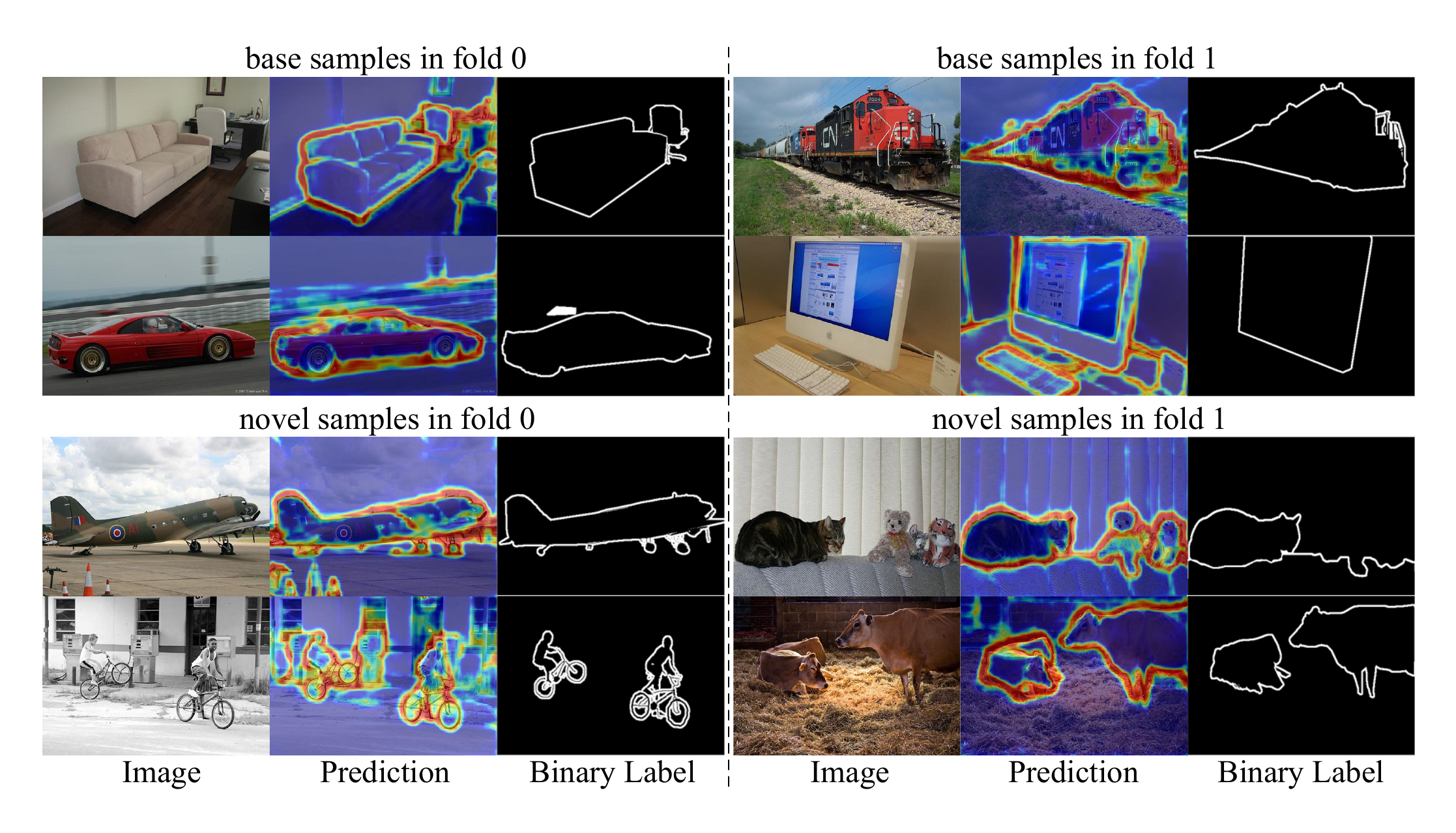}
  \caption{Visualizations on boundary prediction results for base samples and novel samples in fold 0 and fold 1 of PASCAL VOC 2012 \emph{train} set. We display several predicted boundary maps together with their corresponding ground-truth boundary maps.}
  \label{fig:app_boundary}
\end{figure*}

\begin{table*}[t]
  \centering
  \begin{tabular}{cccccccccc}
	\toprule
	& & \multicolumn{4}{c}{PSA} & \multicolumn{4}{c}{RETAB} \\
	\textbf{fold} & sample & acc. & prec. & recall & ${\rm F}_1$ & acc. & prec. & recall & ${\rm F}_1$\\
	\cmidrule(r){3-6}\cmidrule(r){7-10}
	\textbf{0} & base & 0.811 & 0.583 & 0.664 & 0.597 & \textbf{0.841} & \textbf{0.633} & \textbf{0.763} & \textbf{0.662} \\
	\textbf{0} & novel & 0.823 & 0.577 & 0.643 & 0.591 & \textbf{0.862} & \textbf{0.639} & \textbf{0.745} & \textbf{0.669} \\
	\midrule
	\textbf{1} & base & 0.821 & 0.578 & 0.650 & 0.591 & \textbf{0.852} & \textbf{0.634} & \textbf{0.762} & \textbf{0.664} \\
	\textbf{1} & novel & 0.802 & 0.587 & 0.667 & 0.600 & \textbf{0.836} & \textbf{0.628} & \textbf{0.731} & \textbf{0.653} \\
	\bottomrule
  \end{tabular}
  \caption{Analysis on semantic affinities for base samples and novel samples on PASCAL VOC 2012 \emph{train} set. Semantic affinities predicted by PSA and our RETAB are evaluated on fold 0 and fold 1.}
  \label{table:app_affeval}
\end{table*}

\section{Analysis of Boundary Transfer}\label{subsec:app_boundary}

To demonstrate the effectiveness of boundary transfer, we evaluate boundary predictions on PASCAL VOC 2012 \emph{train} set. Figure~\ref{fig:app_boundary} visualizes some boundary maps of base samples and novel samples predicted by our boundary network on fold 0 and fold 1. As discussed in Section 3.2 in the main paper, the predicted boundary map can be transformed into a binary prediction map, where 1 and 0 respectively denote the pixels in boundary regions and non-boundary regions. Ground-truth binary boundary labels for evaluation can be obtained by transforming the semantic boundary labels in SBD~\cite{hariharan2011semantic} into class-agnostic ones. Then we can evaluate the performance of the boundary network in a binary classification manner. As shown in Table~\ref{table:app_boundary}, we respectively calculate the accuracy, precision, recall, and ${\rm F}_1$-score for base samples and novel samples in the \emph{train} set on fold 0 and fold 1. As illustrated, although the boundary network is trained on only base samples, it also performs well in predicting boundaries for novel samples, verifying the effectiveness of knowledge transfer across categories. 

\section{Analysis of Semantic Affinity}\label{subsec:app_affinity}

In this subsection, we compare the semantic affinity learned by PSA~\cite{ahn2018learning} and our RETAB on PASCAL VOC 2012 \emph{train} set. The predicted semantic affinities $\hat{a}_{ij}$ are transformed into the binary values $\hat{a}_{ij}'$ for both base samples and novel samples: $\hat{a}_{ij}'=1$ if $\hat{a}_{ij}\geqslant0.5$, and $\hat{a}_{ij}'=0$ otherwise. Ground-truth semantic affinities $a_{ij}^{*}$ could be obtained by the method described in Section 3.2 in the main paper. Then we can evaluate the predicted semantic affinities within neighbor set $\mathcal{P}$ (see Eqn.(1) in the main paper) in a binary classification manner. We respectively calculate the accuracy, precision, recall, and ${\rm F}_1$-score for base samples and novel samples in the \emph{train} set on fold 0 and fold 1. Table~\ref{table:app_affeval} displays the quantitative results comparing PSA and our RETAB. As illustrated, our method learns more satisfying semantic affinity for both base samples and novel samples than PSA, verifying RETAB's transferability of semantic affinity from base categories to novel categories. 

In Table~\ref{table:app_affinity}, we conduct ablation studies on PASCAL VOC 2012 \emph{train} set to investigate the semantic affinities learned by affinity networks under different training strategies. We show the performance of revised responses for different strategies using the same initial response and the same propagation method. As shown in Table~\ref{table:app_affinity}, using pseudo affinity labels of all samples to train the affinity network (row 1) results in low mIoU scores. When we only utilize ground-truth affinity labels of base samples for training (row 2), the model witnesses a significant performance improvement because more precise semantic affinities can be learned, especially on boundary regions. This phenomenon implies that even if our RETAB (boundary network and affinity network) is only supervised on base samples, affinities could also be transferred well to novel samples. This advantage makes our RETAB more practical in real-world applications because we can directly use a RETAB pretrained on base samples to perform inference on extensive novel samples without further fine-tuning. In row 3 of Table~\ref{table:app_affinity}, we further use pseudo affinities of novel samples to train the affinity network in a mixed-supervised manner, which slightly enhances the performance on novel categories. After filtering out noisy supervisions in boundary regions of CAM for novel samples (row 4), we finally obtain the best strategy used in our RETAB. 

\begin{table*}[t]
  \centering
  \begin{tabular}{cccccccc}
	\toprule
	Supervision for & Supervision for & \multicolumn{3}{c}{fold 0} & \multicolumn{3}{c}{fold 1} \\
	base samples & novel samples & all & base & novel & all & base & novel\\
	\cmidrule(r){1-2}\cmidrule(r){3-5}\cmidrule(r){6-8}
	pseudo & pseudo & 63.6 & 66.4 & 54.8 & 63.6 & 63.1 & 65.2 \\
	GT & - & 70.8 & 73.8 & 61.0 & 70.6 & 70.8 & 69.8 \\
	GT & pseudo & 70.7 & 73.7 & 61.2 & 70.9 & 71.0 & 70.5 \\
	GT & pseudo+filter & \textbf{71.2} & \textbf{74.0} & \textbf{62.5} & \textbf{71.3} & \textbf{71.2} & \textbf{71.6} \\
	\bottomrule
  \end{tabular}
  \caption{The ablation study of semantic affinities learned by affinity networks with different training strategies. Utilizing ``CAM'' as the initial response and our two-stage propagation as the expansion algorithm, we show the mIoU evaluation for revised responses on PASCAL VOC 2012 \emph{train} set. \textbf{GT}: use ground-truth affinity labels obtained from pixel-level segmentation labels. \textbf{pseudo}: use pseudo affinity labels generated from ``CAM''. \textbf{filter}: filter out coordinate pairs in boundary regions.}
  \label{table:app_affinity}
\end{table*}

\begin{table*}[t]
  \centering
  \begin{tabular}{ccccccc}
	\toprule
	\multirow{2}*{Propagation strategy} & \multicolumn{3}{c}{fold 0} & \multicolumn{3}{c}{fold 1} \\
	& all & base & novel & all & base & novel\\
	\cmidrule(r){1-1}\cmidrule(r){2-4}\cmidrule(r){5-7}
	one-stage & 70.7 & 73.7 & 61.1 & 70.3 & 71.1 & 67.9 \\
	nbd+bd & 66.0 & 68.4 & 58.3 & 65.7 & 65.4 & 66.8 \\
	BTP & \textbf{71.2} & \textbf{74.0} & \textbf{62.5} & \textbf{71.3} & \textbf{71.2} & \textbf{71.6} \\
	\bottomrule
  \end{tabular}
  \caption{The ablation study of different propagation strategies on fold 0 and fold 1 of Pascal VOC 2012 \emph{train} set. We show the mIoU evaluation for revised responses, using ``CAM'' as the initial response and the best semantic affinities learned by our RETAB. \textbf{one-stage}: classical random walk. \textbf{nbd+bd}: separately propagate in the non-boundary regions and in the boundary regions. \textbf{BTP}: our proposed boundary-aware two-stage propagation strategy.}
  \label{table:app_btp}
\end{table*}

\begin{table*}[t]
  \centering
  \begin{tabular}{cccccccc}
	\toprule
	WSSS & affgt & FILTER & BTP & seggt & all-mIoU & base-mIoU & novel-mIoU \\
	\midrule
	$\checkmark$ &&&&& 61.7 & 62.8 & 58.1 \\
	$\checkmark$ &&&& $\checkmark$ & 71.9 & 76.3 & 57.8 \\
	$\checkmark$ & $\checkmark$ &&& $\checkmark$ & 73.9 & 76.6 & 65.1 \\
	$\checkmark$ & $\checkmark$ & $\checkmark$ && $\checkmark$ & 74.3 & 76.6 & 67.0 \\
	$\checkmark$ & $\checkmark$ & $\checkmark$ & $\checkmark$ & $\checkmark$ & \textbf{75.0} & \textbf{76.8} & \textbf{69.2} \\
	\bottomrule
  \end{tabular}
  \caption{The ablation study for each source of improvement in RETAB. Models are trained on fold 0 and the final segmentation results are evaluated on VOC12 \emph{val} set. \textbf{WSSS}: weakly-supervised semantic segmentation baseline PSA~\cite{ahn2018learning}. \textbf{affgt}: use ground-truth affinity labels for base samples when training the affinity network. \textbf{FILTER}: filter out coordinate pairs in boundary regions for novel samples. \textbf{BTP}: boundary-aware two-stage propagation. \textbf{seggt}: use ground-truth segmentation labels for base samples when training the segmentation network.}
  \label{table:improve}
\end{table*}

\section{Analysis of Boundary-aware Two-stage Propagation}\label{subsec:app_propagation}

By taking fold 0 and fold 1 as examples, we analyze the functionality of boundary-aware two-stage propagation on Pascal VOC 2012 \emph{train} set. We compare three propagation methods: 1) ``one-stage'': classical one-stage random walk~\cite{ahn2018learning}, 2) ``nbd+bd'': separately propagate in the non-boundary region and in the boundary region, where confident pixels can not guide the propagation of unconfident ones, 3) ``BTP'': our boundary-aware two-stage propagation as described in Section 3.2 in the main paper. We adopt CAM~\cite{zhou2016learning} as the initial response and use the best semantic affinities learned by our RETAB. The results are summarized in Table~\ref{table:app_btp}. As illustrated, our proposed ``BTP'' outperforms both ``one-stage'' and ``nbd+bd'' especially on novel categories, verifying the importance of utilizing confident pixels to guide the propagation of unconfident ones. 

\section{Source of Improvement}\label{subsec:app_improve}

Table~\ref{table:improve} gives an ablation study of each part in our approach that contributes to the improvement of the final segmentation results on PASCAL VOC 2012 \emph{val} set. Experiments are conducted on fold 0. In Table~\ref{table:improve}, row 1 is the WSSS baseline PSA (CAM+RW)~\cite{ahn2018learning}. Row 2 is our created naive weak-shot segmentation baseline ``seggt'' and row 3 is the augmented weak-shot segmentation baseline ``affgt+seggt''. Except row 1, all experiments adopt ground-truth segmentation masks of base samples so as to train the segmentation network in a mixed-supervised manner. Comparing row 3 with row 2, we find that applying ground-truth affinity labels of base samples to train the affinity network can improve $0.3$ base-mIoU ($\%$) and $7.3$ novel-mIoU ($\%$). After filtering out noisy affinity labels in novel samples, the model in row 4 gains another $1.9$ improvement on novel-mIoU ($\%$). Comparing row 5 with row 4, replacing the original random walk with our proposed boundary-aware two-stage propagation further enhances $2.2$ novel-mIoU ($\%$), resulting in RETAB with the best mIoUs.

\begin{table*}[!t]
	\centering
	\begin{tabular}{p{7mm}<{\centering}p{4.5mm}<{\centering}p{4.5mm}<{\centering}p{7mm}<{\centering}p{4.5mm}<{\centering}p{4.5mm}<{\centering}p{7mm}<{\centering}p{4.5mm}<{\centering}p{4.5mm}<{\centering}p{7mm}<{\centering}p{4.5mm}<{\centering}p{4.5mm}<{\centering}p{7mm}<{\centering}}
		\toprule
		& \multicolumn{3}{c}{CAM} & \multicolumn{3}{c}{CAM+RW} & \multicolumn{3}{c}{CAM+RW(affgt)} & \multicolumn{3}{c}{CAM+RETAB} \\
		\textbf{fold} & all & base & novel & all & base & novel & all & base & novel & all & base & novel \\
		\cmidrule(r){2-4}\cmidrule(r){5-7}\cmidrule(r){8-10}\cmidrule(r){11-13}
		\textbf{0} & 48.0 & 51.4 & 37.4 & 61.0 & 63.8 & 52.3 & 69.7 & 73.1 & 58.7 & \textbf{71.2} & \textbf{74.0} & \textbf{62.5} \\
		\textbf{1} & 48.0 & 47.8 & 48.8 & 61.0 & 61.0 & 61.2 & 70.0 & 71.0 & 66.6 & \textbf{71.3} & \textbf{71.2} & \textbf{71.6} \\
		\textbf{2} & 48.0 & 47.2 & 50.7 & 61.0 & 58.9 & 67.9 & 69.7 & 69.6 & 70.0 & \textbf{70.9} & \textbf{70.2} & \textbf{73.3} \\
		\textbf{3} & 48.0 & 47.6 & 49.4 & 61.0 & 62.0 & 57.9 & 69.0 & 71.9 & 59.8 & \textbf{70.1} & \textbf{72.4} & \textbf{62.8} \\
		\midrule
		& \multicolumn{3}{c}{SEAM} & \multicolumn{3}{c}{SEAM+RW} & \multicolumn{3}{c}{SEAM+RW(affgt)} & \multicolumn{3}{c}{SEAM+RETAB} \\
		\textbf{fold} & all & base & novel & all & base & novel & all & base & novel & all & base & novel \\
		\cmidrule(r){2-4}\cmidrule(r){5-7}\cmidrule(r){8-10}\cmidrule(r){11-13}
		\textbf{0} & 55.4 & 58.8 & 44.6 & 63.6 & 66.4 & 54.8 & 67.5 & 71.1 & 56.1 & \textbf{68.2} & \textbf{71.2} & \textbf{58.4} \\
		\textbf{1} & 55.4 & 55.3 & 55.7 & 63.6 & 63.1 & 65.2 & 67.2 & 67.1 & 67.7 & \textbf{68.1} & \textbf{67.6} & \textbf{69.7} \\
		\textbf{2} & 55.4 & 53.2 & 62.3 & 63.6 & 60.9 & 72.4 & 67.0 & 65.1 & 73.2 & \textbf{67.9} & \textbf{65.7} & \textbf{74.8} \\
		\textbf{3} & 55.4 & 56.0 & 53.6 & 63.6 & 65.5 & 57.6 & 66.7 & 69.4 & 58.2 & \textbf{67.5} & \textbf{69.7} & \textbf{60.7} \\
		\midrule
		& \multicolumn{3}{c}{CPN} & \multicolumn{3}{c}{CPN+RW} & \multicolumn{3}{c}{CPN+RW(affgt)} & \multicolumn{3}{c}{CPN+RETAB} \\
		\textbf{fold} & all & base & novel & all & base & novel & all & base & novel & all & base & novel \\
		\cmidrule(r){2-4}\cmidrule(r){5-7}\cmidrule(r){8-10}\cmidrule(r){11-13}
		\textbf{0} & 57.4 & 60.5 & 47.5 & 67.8 & 70.5 & 59.3 & 70.5 & 73.0 & 62.3 & \textbf{71.5} & \textbf{73.3} & \textbf{65.9} \\
		\textbf{1} & 57.4 & 55.9 & 62.3 & 67.8 & 66.6 & 71.6 & 71.2 & 70.7 & 72.9 & \textbf{72.1} & \textbf{71.3} & \textbf{74.6} \\
		\textbf{2} & 57.4 & 56.3 & 61.0 & 67.8 & 66.3 & 72.7 & 71.6 & 70.9 & 74.0 & \textbf{72.3} & \textbf{71.5} & \textbf{75.0} \\
		\textbf{3} & 57.4 & 58.6 & 53.8 & 67.8 & 69.3 & 62.9 & 70.2 & 72.2 & 64.0 & \textbf{70.7} & \textbf{72.4} & \textbf{65.1} \\
		\bottomrule
	\end{tabular}
	\caption{Mean IoU evaluation of initial responses and revised responses on VOC12 \emph{train} set. We adopt three initial responses, \emph{i.e.}, ``CAM''~\cite{zhou2016learning}, ``SEAM''~\cite{wang2020self}, and ``CPN''~\cite{zhang2021complementary}. We compare three response propagation methods: ``RW'', ``RW(affgt)'', and ``RETAB''. ``RW'' is introduced in PSA~\cite{ahn2018learning}. ``RW(affgt)'' is based on ``RW'' and further uses ground-truth affinity labels of base samples when training the affinity network. ``RETAB'' is our proposed method.}
	\label{table:cam}
\end{table*}

\begin{table*}[t]
	\centering
	\begin{tabular}{p{32mm}p{3.5mm}<{\centering}p{3.5mm}<{\centering}p{3.5mm}<{\centering}p{3.5mm}<{\centering}p{3.5mm}<{\centering}p{3.5mm}<{\centering}p{3.5mm}<{\centering}p{3.5mm}<{\centering}p{3.5mm}<{\centering}p{3.5mm}<{\centering}p{3.5mm}<{\centering}p{3.5mm}<{\centering}}
		\toprule
		\multirow{2}*{Method} & \multicolumn{3}{c}{fold 0} & \multicolumn{3}{c}{fold 1} & \multicolumn{3}{c}{fold 2} & \multicolumn{3}{c}{fold 3} \\
		& $\mathcal{C}$ & $\mathcal{C}^b$ & $\mathcal{C}^n$ & $\mathcal{C}$ & $\mathcal{C}^b$ & $\mathcal{C}^n$ & $\mathcal{C}$ & $\mathcal{C}^b$ & $\mathcal{C}^n$ & $\mathcal{C}$ & $\mathcal{C}^b$ & $\mathcal{C}^n$ \\
		\midrule
		SSDD~\cite{shimoda2019self} & 64.9 & 67.2 & 57.5 & 64.9 & 62.9 & 71.2 & 64.9 & 63.2 & 70.4 & 64.9 & 67.8 & 55.7 \\
		BES~\cite{chen2020weakly} & 65.7 & 67.0 & 61.7 & 65.7 & 63.8 & 71.6 & 65.7 & 64.7 & 68.8 & 65.7 & 68.7 & 56.0 \\
		SvM~\cite{zhang2020splitting} & 66.6 & 66.3 & 67.8 & 66.6 & 65.9 & 68.9 & 66.6 & 66.2 & 68.1 & 66.6 & 69.6 & 57.0 \\
		\midrule
		CAM+RW~\cite{ahn2018learning} & 61.7 & 62.8 & 58.1 & 61.7 & 61.3 & 62.8 & 61.7 & 60.5 & 65.3 & 61.7 & 63.7 & 55.1 \\
		CAM+RW(seggt) & 71.9 & 76.3 & 57.8 & 73.7 & 74.9 & 69.9 & 72.9 & 74.1 & 69.0 & 72.9 & 78.3 & 55.8 \\
		CAM+RW(affgt+seggt) & 73.9 & 76.6 & 65.1 & 74.6 & 75.0 & 73.3 & 73.6 & 74.3 & 71.4 & 73.2 & \textbf{78.4} & 56.6 \\
		CAM+RETAB & \textbf{75.0} & \textbf{76.8} & \textbf{69.2} & \textbf{75.7} & \textbf{75.6} & \textbf{76.1} & \textbf{74.2} & \textbf{74.9} & \textbf{72.0} & \textbf{73.7} & \textbf{78.4} & \textbf{58.5} \\
		\midrule
		SEAM+RW~\cite{wang2020self} & 64.5 & 66.2 & 59.0 & 64.5 & 63.3 & 68.3 & 64.5 & 62.9 & 69.7 & 64.5 & 67.1 & 56.2 \\
		SEAM+RW(seggt) & 72.4 & 76.2 & 60.2 & 74.1 & 74.5 & 72.6 & 72.6 & 72.9 & 71.7 & 73.2 & 78.2 & 57.4 \\
		SEAM+RW(affgt+seggt) & 73.3 & 76.5 & 63.1 & 74.6 & 74.7 & 74.3 & 73.1 & 73.3 & 72.5 & 73.8 & 78.7 & 58.1 \\
		SEAM+RETAB & \textbf{74.0} & \textbf{76.6} & \textbf{65.4} & \textbf{74.9} & \textbf{75.0} & \textbf{74.5} & \textbf{73.6} & \textbf{73.7} & \textbf{73.0} & \textbf{74.3} & \textbf{79.1} & \textbf{58.9} \\
		\midrule
		CPN+RW~\cite{zhang2021complementary} & 67.8 & 69.1 & 63.8 & 67.8 & 65.9 & 73.7 & 67.8 & 66.8 & 70.9 & 67.8 & 70.8 & 58.1 \\
		CPN+RW(seggt) & 73.7 & 76.4 & 65.0 & 74.9 & 75.0 & 74.6 & 74.0 & 74.9 & 71.0 & 73.5 & 78.1 & 58.8 \\
		CPN+RW(affgt+seggt) & 74.7 & \textbf{76.8} & 68.1 & 75.4 & 75.5 & 75.2 & 74.3 & 75.0 & 71.9 & 73.9 & 78.6 & 59.0 \\
		CPN+RETAB & \textbf{75.1} & 76.7 & \textbf{70.0} & \textbf{75.9} & \textbf{75.7} & \textbf{76.5} & \textbf{74.8} & \textbf{75.3} & \textbf{73.3} & \textbf{74.5} & \textbf{79.0} & \textbf{60.3} \\
		\midrule
		Fully Oracle & 76.6 & 76.8 & 76.1 & 76.6 & 76.2 & 78.1 & 76.6 & 75.4 & 80.6 & 76.6 & 79.2 & 68.2 \\
		\bottomrule
	\end{tabular}
	\caption{Comparison of segmentation performance on PASCAL VOC 2012 \emph{val} set. Columns marked by $\mathcal{C}$, $\mathcal{C}^b$ and $\mathcal{C}^n$ represents all-mIoU, base-mIoU and novel-mIoU, respectively. We have five groups of methods. The first group includes some popular WSSS methods. The second / third / fourth group represents our WSSS baseline PSA / SEAM / CPN, together with our created weak-shot segmentation baselines (seggt and affgt+seggt), and our implemented RETAB with the same initial response. The fifth group is a fully-supervised method. Except the first group, all experiments use DeepLab~\cite{chen2014semantic} with ResNet38~\cite{wu2019wider} backbone.}
	\label{table:app_seg}
\end{table*}



\section{Evaluation on Pseudo Segmentation Labels}\label{subsec:app_pseudores}

We evaluate pseudo segmentation labels by assessing CAMs before/after propagation on PASCAL VOC 2012 \emph{train} set. To verify the transferability of our method, we adopt three initial responses: ``CAM'', ``SEAM'' and ``CPN''. Table~\ref{table:cam} compares the performance of three response expanding methods, \emph{i.e.}, ``RW'', ``RW(affgt)'' and our proposed ``RETAB'', where ``RW(affgt)'' is based on ``RW'' and further uses ground-truth affinity labels of base samples when training the affinity network. As illustrated, ``RW(affgt)'' greatly advances base-mIoUs since more accurate semantic affinities in base samples are learned. Still, this method can not achieve satisfying results on novel categories. By comparison, our ``RETAB'' achieves consistently higher mIoU results than baselines on all four folds. In detail, ``RETAB'' improves $9.1\sim10.3$ all-mIoU ($\%$), $10.2\sim11.3$ base-mIoU ($\%$) and $4.9\sim10.4$ novel-mIoU ($\%$) when compared with ``RW'' in PSA\footnote{The official \emph{train} set all-mIoU of revised responses is $58.1$ in PSA~\cite{ahn2018learning}, and our reproduced result is $61.0$.}, improves $3.9\sim4.6$ all-mIoU ($\%$), $4.2\sim4.8$ base-mIoU ($\%$) and $2.4\sim4.5$ novel-mIoU ($\%$) when compared with ``RW'' in SEAM, and improves $4.3\sim6.6$ all-mIoU ($\%$), $4.0\sim7.8$ base-mIoU ($\%$) and $3.2\sim11.1$ novel-mIoU ($\%$) when compared with ``RW'' in CPN. These improvements come from two aspects. On the one hand, we utilize more confident semantic affinities by incorporating ground-truth labels and boundary predictions in affinity learning. On the other hand, two-stage propagation makes it possible that confident pixels guide the propagation of unconfident ones. Interestingly, although the initial response ``SEAM'' outperforms ``CAM'', ``CAM+RETAB'' works better than ``SEAM+RETAB'' after propagation. It shows that the usefulness of RETAB is related to the chosen initial response.

\section{Evaluation on Segmentation Performance}\label{subsec:app_segres}

This section supplements Section 4.3 in the main paper. Table 1 in the main paper shows the segmentation performance on the \emph{test} set. In this section, Table~\ref{table:app_seg} further provides results on the \emph{val} set. By comparison, our method still outperforms baselines on all folds, especially for novel categories, which is consistent with the performance on the \emph{test} set.

\begin{figure*}[!ht]
  \centering
  \includegraphics[width=\linewidth]{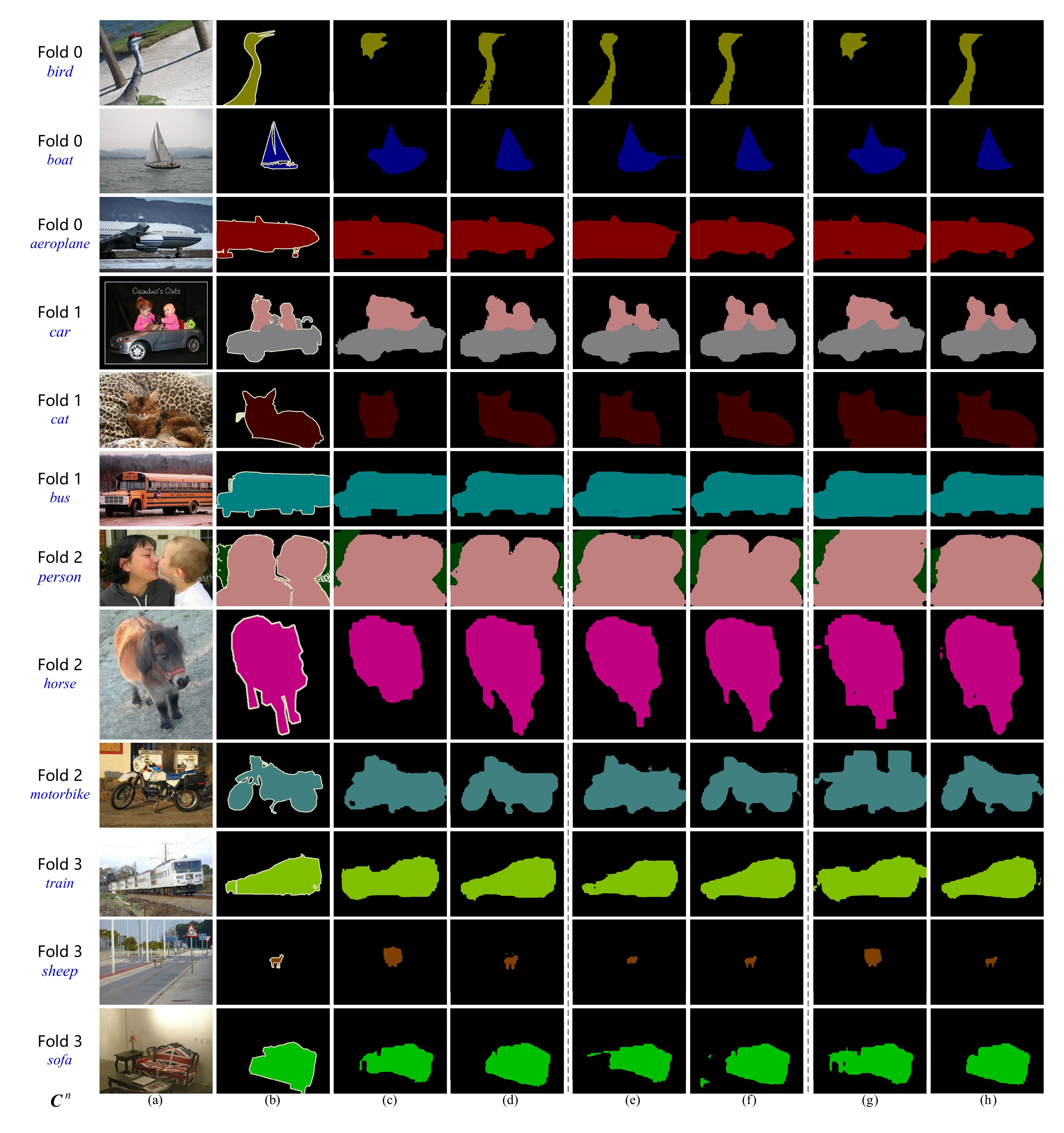}
  \caption{Visualized pseudo segmentation labels of novel samples on PASCAL VOC 2012 \emph{train} set. (a) image. (b) GT label. (c) ``CAM+RW''. (d) ``CAM+RETAB''. (e) ``SEAM+RW''. (f) ``SEAM+RETAB''. (g) ``CPN+RW''. (h) ``CPN+RETAB''. Names of novel categories presented in the images are marked as blue on the left side.}
  \label{fig:app_pseudo}
\end{figure*}

\begin{figure*}[!ht]
  \centering
  \includegraphics[width=\linewidth]{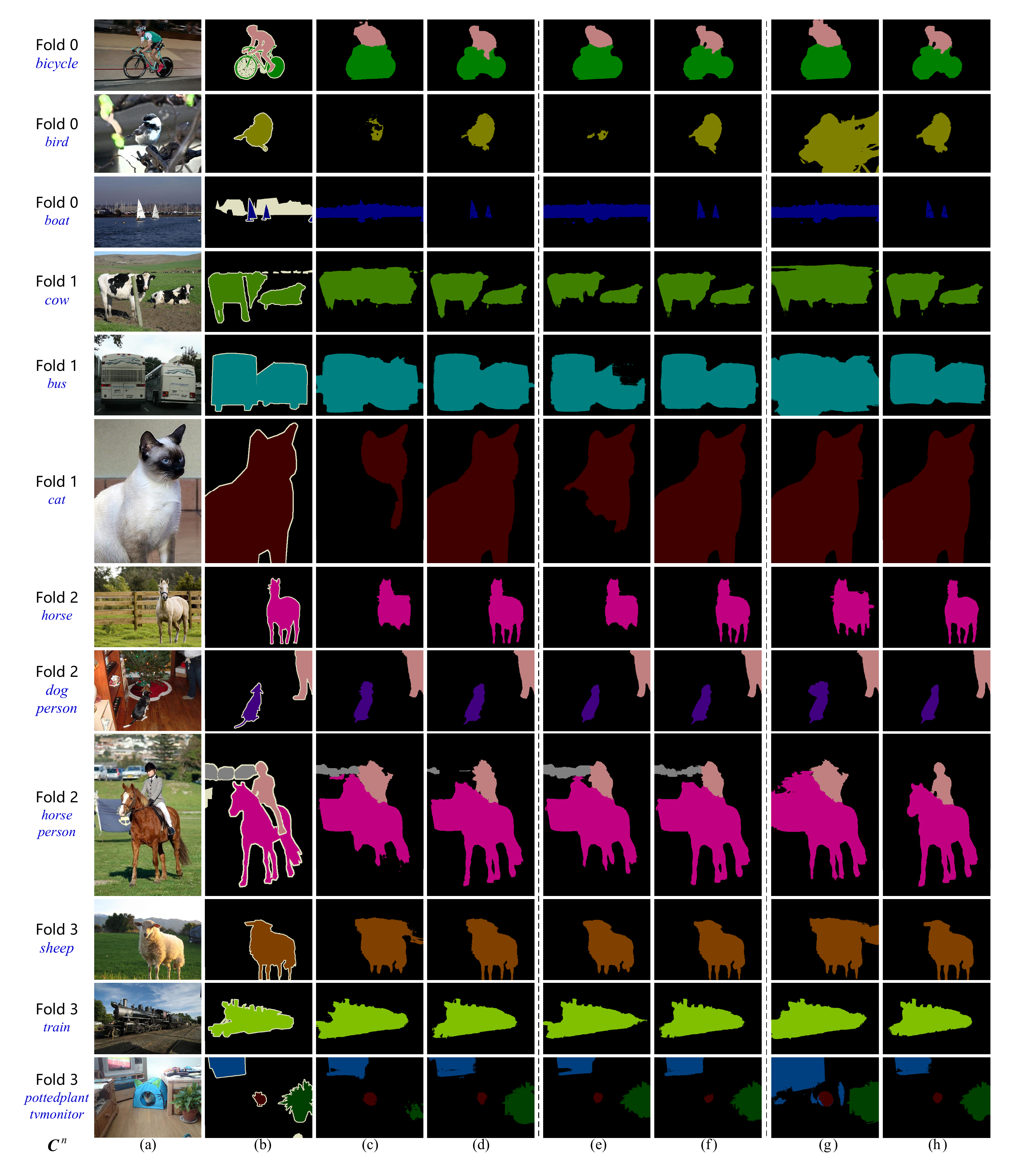}
  \caption{Qualitative semantic segmentation results for novel categories on PASCAL VOC 2012 \emph{val} set. (a) image. (b) GT label. (c) ``CAM+RW''. (d) ``CAM+RETAB''. (e) ``SEAM+RW''. (f) ``SEAM+RETAB''. (g) ``CPN+RW''. (h) ``CPN+RETAB''. Names of novel categories presented in the images are marked as blue on the left side.}
  \label{fig:app_seg}
\end{figure*}

\section{Qualitative Results on Pseudo Segmentation Labels}\label{subsec:app_vispseudo}

In Figure~\ref{fig:app_pseudo}, we show more visualizations of pseudo segmentation labels for novel samples generated by different methods in fold 0, 1, 2, and 3 on PASCAL VOC 2012 \emph{train} set, which supplements Figure 3 in the main paper. In Figure~\ref{fig:app_pseudo}, column (c), (e), and (g) represent the WSSS baseline PSA~\cite{ahn2018learning}, SEAM~\cite{wang2020self}, and CPN~\cite{zhang2021complementary}. Column (d), (f), and (h) denote our implemented RETAB with the same initial responses as these three baselines, respectively. By comparing (c) with (d), comparing (e) with (f), and comparing (g) with (h), we find that our method outperforms all baselines in recovering object shapes on the boundary regions of pseudo labels. Boundary-aware two-stage propagation restricts the random walk process within the target object regions and allows confident pixels to guide unconfident ones, resulting in pseudo segmentation labels with higher quality.

\section{Qualitative Results on Semantic Segmentation}\label{subsec:app_visseg}

We show more visualizations in Figure~\ref{fig:app_seg} for the segmentation results on PASCAL VOC 2012 \emph{val} set predicted by different methods in fold 0, 1, 2 and 3, which supplements Figure 4 in the main paper. In Figure~\ref{fig:app_seg}, column (c), (e), and (g) respectively displays the WSSS baseline PSA~\cite{ahn2018learning}, SEAM~\cite{wang2020self}, and CPN~\cite{zhang2021complementary}. It is obvious that object boundaries predicted by these baselines are not precise. Also, some adjacent objects are wrongly classified as inaccurate categories. Column (d), (f), and (h) in Figure~\ref{fig:app_seg} represent our implemented RETAB with the same initial responses and the same segmentation network as three baselines, respectively. For a fair comparison, we compare (c) with (d), compare (e) with (f), and compare (g) and (h) to verify the effectiveness of our method. As illustrated, RETAB performs considerably better than three baselines in filling object regions while not overstepping the object boundaries. Also, our method can output more delicate details, like animal legs and plant leaves. Consequently, we can safely claim that RETAB successfully transfers pixel-level knowledge from base categories to novel ones and narrows the gap between weakly-supervised and fully-supervised segmentation.

\section{Generalization to Potential Novel Categories in the Background}\label{subsec:app_potential}

Recall that the basic goal of weak-shot semantic segmentation is to improve the segmentation performance on weakly-annotated images (novel samples) with an auxiliary fully-annotated dataset (base samples). In practical scenarios, base samples might contain potential novel categories in some regions labeled ``background''. Specifically, when labeling the fully-annotated dataset, certain categories (base categories) are selected to annotate precisely, and all the other pixels are marked as ``background''. Actually, these background pixels may contain novel categories we aim at. If we can discover these potential novel categories in base samples, we might achieve better segmentation performances on novel categories for the newly-coming weakly-annotated images.

To simulate the above situation, we propose two novel folds for PASCAL VOC 2012  dataset: fold 0$^*$ and fold 1$^*$. They follow the same category split rule as fold 0 and fold 1, respectively. Furthermore, based on the split of base samples and novel samples in fold 0 and fold 1, we further randomly choose approximately $50\%$ novel samples and move them to base samples by changing their pixel labels of novel categories into the ``background'' label with index 0. We refer to these new labels as revised segmentation labels. Consequently, base samples now contain potential novel categories in the background. After this operation, the number of base samples increases, and the number of novel samples decreases. In our experiments, fold 0$^*$ contains 9184 base / 1398 novel samples, and fold 1$^*$ contains 8755 base / 1827 novel samples in the $trainaug$ set. Next, we discuss the training strategy to deal with this typical problem, which is also included in our weak-shot semantic segmentation task.

\begin{table*}[t]
	\centering
	\begin{tabular}{lcccccc}
		\toprule
		\multirow{2}*{Method} & \multicolumn{3}{c}{fold 0$^*$} & \multicolumn{3}{c}{fold 1$^*$} \\
		& $\mathcal{C}$ & $\mathcal{C}^b$ & $\mathcal{C}^n$ & $\mathcal{C}$ & $\mathcal{C}^b$ & $\mathcal{C}^n$ \\
		\midrule
		CAM+RW(seggt) & 66.3 & 76.0 & 35.1 & 63.0 & 74.8 & 25.1 \\
		CAM+RW(affgt+seggt) & 67.5 & 76.1 & 39.9 & 64.6 & 75.0 & 31.3 \\
		CAM+RETAB & 67.8 & 74.4 & 46.5 & 65.3 & 74.4 & 36.3 \\
		CAM+RETAB+ST & \textbf{71.9} & \textbf{76.2} & \textbf{58.0} & \textbf{74.5} & \textbf{75.8} & \textbf{70.3} \\
		\midrule
		Fully Oracle & 76.6 & 76.8 & 76.1 & 76.6 & 76.2 & 78.1 \\
		\bottomrule
	\end{tabular}
	\caption{Comparison of segmentation performance on PASCAL VOC 2012 \emph{val} set for fold 0$^*$ and fold 1$^*$. ``CAM+RW(seggt)'' and ``CAM+RW(affgt+seggt)'' are weak-shot segmentation baselines. ``CAM+RETAB'' denotes our implemented RETAB. ``CAM+RETAB+ST'' denotes our method followed by a self-training step. ``Fully Oracle'' is the fully-supervised upper bound. All experiments use DeepLab~\cite{chen2014semantic} with ResNet38~\cite{wu2019wider} backbone for segmentation.}
	\label{table:app_potential}
\end{table*}

Based on the new split and the revised segmentation labels, the training process of our RETAB and the segmentation network remains unchanged. Furthermore, we adopt a self-training strategy (\emph{abbr.} ST) after the segmentation step to discover potential novel categories in the background of base samples. ST can be regarded as a self-supervised fine-tuning step for the segmentation network with mixed supervisions from base samples and novel samples. We introduce each iteration of ST in detail as follows, which includes three steps. First, we use the current segmentation network to perform segmentation predictions on the current batch of training samples. Second, for the background pixels classified as novel categories, we flip the pixel label from ``background'' to the predicted novel category for each of these pixels. Third, the revised segmentation labels are used to fine-tune the segmentation network via back-propagation. By gradually discovering potential novel categories in the background, the segmentation performance of novel categories could be iteratively improved. The above procedure is repeated until a predefined number of iteration is reached. In our implementation, we apply ST for 10000 iterations. Except that the initial learning rate is adjusted to half, all other parameters are not changed when we fine-tune the segmentation network.

Table~\ref{table:app_potential} summarizes segmentation results on the \emph{val} set for fold 0$^*$ and fold 1$^*$. We list four methods: 1) our created naive weak-shot baseline ``CAM+RW(seggt)'' which uses revised labels of base samples and pseudo labels of novel samples generated by PSA~\cite{ahn2018learning} for segmentation, 2) our created augmented weak-shot baseline ``CAM+RW(affgt+seggt)'' which is based on the naive baseline and further uses ground-truth affinity labels of base samples for affinity training, 3) our method ``CAM+RETAB'' which uses revised labels of base samples and pseudo labels of novel samples generated by our RETAB for segmentation, and 4) our method with self-training ``CAM+RETAB+ST''. As shown in Table~\ref{table:app_potential}, our method consistently outperforms all the baselines on novel categories. After adopting the self-training strategy, more potential novel categories in the base samples are discovered. As a result, the segmentation network witnesses a huge improvement on novel categories. In conclusion, our RETAB could generalize to discover potential cues of novel categories in the background of base samples with a simple follow-up self-training step.

\begin{table*}[t]
  \scriptsize
  \centering
  \begin{tabular}{p{5.8mm}p{0mm}p{1mm}p{1.2mm}p{1.2mm}p{1.2mm}p{2.4mm}p{0.3mm}p{0.2mm}p{0mm}p{1.8mm}p{0.9mm}p{1.8mm}p{0.6mm}p{2.1mm}p{1.3mm}p{3.4mm}p{2.3mm}p{2.5mm}p{1.7mm}p{2.1mm}p{1.4mm}}
	\toprule
	\textbf{class} & bg & aero & bike & bird & boat & bottle & bus & car & cat & chair & cow & table & dog & horse & mbk & person & plant & sheep & sofa & train & tv \\
	\textbf{index} & 0 & 1 & 2 & 3 & 4 & 5 & 6 & 7 & 8 & 9 & 10 & 11 & 12 & 13 & 14 & 15 & 16 & 17 & 18 & 19 & 20 \\
	\midrule
    \textbf{fold 4} & $b$ & $n$ & $n$ & $n$ & $n$ & $n$ & $n$ & $n$ & $n$ & $n$ & $n$ & $b$ & $b$ & $b$ & $b$ & $b$ & $b$ & $b$ & $b$ & $b$ & $b$ \\
    \textbf{fold 5} & $b$ & $n$ & $n$ & $n$ & $n$ & $n$ & $n$ & $n$ & $n$ & $n$ & $n$ & $n$ & $n$ & $n$ & $n$ & $n$ & $b$ & $b$ & $b$ & $b$ & $b$ \\
	\bottomrule
  \end{tabular}
  \caption{The category splits of fold 4 and fold 5 for PASCAL VOC 2012 dateset. In each fold, `$b$' denotes a base category and `$n$' denotes a novel category. $bg$ represents for the background category.}
  \label{table:app_extendeddataset}
\end{table*}

\begin{table*}[t]
	\centering
	\begin{tabular}{lcccccc}
		\toprule
		\multirow{2}*{Method} & \multicolumn{3}{c}{fold 4} & \multicolumn{3}{c}{fold 5} \\
		& $\mathcal{C}$ & $\mathcal{C}^b$ & $\mathcal{C}^n$ & $\mathcal{C}$ & $\mathcal{C}^b$ & $\mathcal{C}^n$ \\
		\midrule
		CAM+RW~\cite{ahn2018learning} & 61.7 & 62.8 & 60.5 & 61.7 & 60.6 & 62.1 \\
		CAM+RW(seggt) & 67.7 & 72.6 & 62.4 & 64.2 & 67.8 & 62.8 \\
		CAM+RW(affgt+seggt) & 70.6 & 74.8 & 66.9 & 66.7 & 70.9 & 65.0 \\
		CAM+RETAB & \textbf{74.3} & \textbf{75.6} & \textbf{72.8} & \textbf{69.3} & \textbf{71.5} & \textbf{68.4} \\
		\midrule
		Fully Oracle & 76.6 & 76.2 & 77.1 & 76.6 & 72.6 & 78.2 \\
		\bottomrule
	\end{tabular}
	\caption{Comparison of segmentation performance on fold 4 and fold 5 of PASCAL VOC 2012 \emph{val} set. ``CAM+RW'' represents our WSSS baseline PSA. ``CAM+RW(seggt)'' and ``CAM+RW(affgt+seggt)'' are weak-shot segmentation baselines. ``CAM+RETAB'' denotes our implemented RETAB. ``Fully Oracle'' is the fully-supervised upper bound. All experiments use DeepLab~\cite{chen2014semantic} with ResNet38~\cite{wu2019wider} backbone for segmentation.}
	\label{table:app_unbalanced}
\end{table*}

\section{Generalization to Fewer Fully-annotated Training Data}\label{subsec:app_unbalanced}

To prove that RETAB can use a small proportion of base samples to facilitate a large number of novel samples, we create fold 4 and fold 5 to include more novel categories and fewer base categories. As shown in Table~\ref{table:app_extendeddataset}, there are 10 novel categories in fold 4 and 15 novel categories in fold 5. Fold 4 contains 4443 base / 6139 novel samples, and fold 5 contains 1014 base / 9568 novel samples in the $trainaug$ set. Following the evaluation setting in Section 4.3 in the main paper, we compare our method ``CAM+RETAB'' with the WSSS baseline ``CAM+RW'' and our created weak-shot segmentation baselines: ``CAM+RW(seggt)'' and  ``CAM+RW(affgt+seggt)''. The segmentation results on the \emph{val} set are listed in Table~\ref{table:app_unbalanced}. Our method surpasses all baseline methods, especially on novel categories. Also, it can recover a considerable proportion of the fully-supervised upper bound. These evaluation results prove that our proposed RETAB can generalize well even if weakly-annotated novel samples are much more than fully-annotated base samples.

\section{Limitations and Future Works}\label{subsec:app_limitation}

\begin{figure*}[t]
  \centering
  \includegraphics[width=0.94\linewidth]{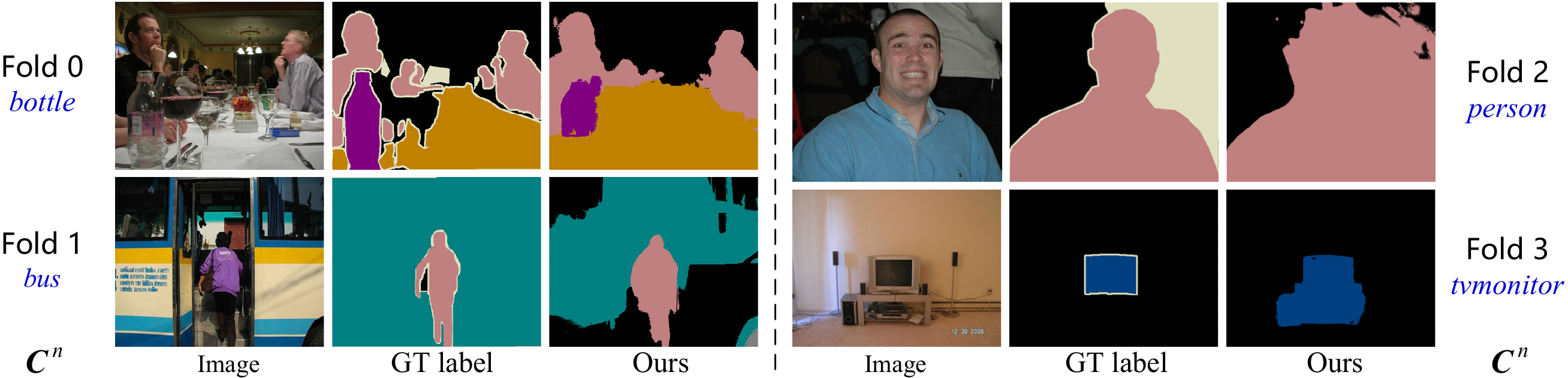}
  \caption{Failure cases of semantic segmentation results on VOC12 \emph{val} set. Names of novel categories presented in the images are marked as blue on both sides of the figure.}
  \label{fig:app_failure}
\end{figure*}

In Figure~\ref{fig:app_failure}, we list some failure cases of segmentation results of our method on the \emph{val} set. Although RETAB successfully narrows the gap between weakly-supervised and fully-supervised segmentation, it could also probably make unsatisfying predictions on some hard novel categories, like ``bottle'' in fold 0 and ``person'' in fold 2. ``bottle'' objects tend to have paster on the body, which could easily be treated as part of the boundaries, thus misleading the propagation to undesirable results. ``person'' objects take up a considerable proportion in the dataset. This category has spatial relationships with many other categories, making ``person'' hard to segment when it is split into the set of novel categories.

In this work, we focus on the second step under the typical WSSS framework to improve the response expansion algorithm. However, the first step, \emph{i.e.}, the way to obtain the initial response, is also crucial. The quality of the initial response directly affects the performance of downstream operations. In the future, we will focus on how to transfer pixel-level knowledge from base categories to novel categories to improve the performance of the initial response for novel samples. Since our proposed RETAB can work on any initial response, the improved initial response could also use RETAB as the second step for response expansion. We hope that RETAB can function as a simple yet effective baseline to facilitate future researches on weak-shot semantic segmentation.

\bibliography{ref}